\def\year{2021}\relax
\documentclass[letterpaper]{article}
\usepackage{aaai21}
\usepackage{times}
\usepackage{helvet}
\usepackage{courier}
\usepackage[hyphens]{url}
\usepackage{graphicx}
\usepackage{natbib}
\usepackage{caption}
\usepackage{amsmath}
\usepackage[algo2e,noend,boxed]{algorithm2e}
\usepackage{algorithmic}
\usepackage{algorithm}
\usepackage{booktabs}

\usepackage[switch]{lineno}

\title{Holistic Multi-View Building Analysis in the Wild with Projection Pooling}

\author {
    Zbigniew Wojna,\textsuperscript{\rm 1}
    Krzysztof Maziarz,\thanks{Work done during an internship at Tensorflight. The author is now at Microsoft Research.}\textsuperscript{\rm 2}
    {\L}ukasz Jocz,\textsuperscript{\rm 1}\\
    Robert Pa{\l}uba,\textsuperscript{\rm 1}
    Robert Kozikowski,\textsuperscript{\rm 1}
    Iasonas Kokkinos\textsuperscript{\rm 3}\\
}
\affiliations {
    \textsuperscript{\rm 1} Tensorflight\\
    \textsuperscript{\rm 2} Jagiellonian University\\
    \textsuperscript{\rm 3} University College London\\
}

\begin{document}

\maketitle

\begin{abstract}
We address six different classification tasks related to fine-grained building attributes: construction type, number of floors, pitch and geometry of the roof, facade material, and occupancy class. Tackling such a remote building analysis problem became possible only recently due to growing large-scale datasets of urban scenes. To this end, we introduce a new benchmarking dataset, consisting of 49426 images (top-view and street-view) of 9674 buildings. These photos are further assembled, together with the geometric metadata. The dataset showcases various real-world challenges, such as occlusions, blur, partially visible objects, and a broad spectrum of buildings. We propose a new \emph{projection pooling layer}, creating a unified, top-view representation of the top-view and the side views in a high-dimensional space. It allows us to utilize the building and imagery metadata seamlessly. Introducing this layer improves classification accuracy -- compared to highly tuned baseline models -- indicating its suitability for building analysis.
\end{abstract}

\section{Introduction}
\begin{figure*}
\centering
\includegraphics[width=0.85\linewidth]{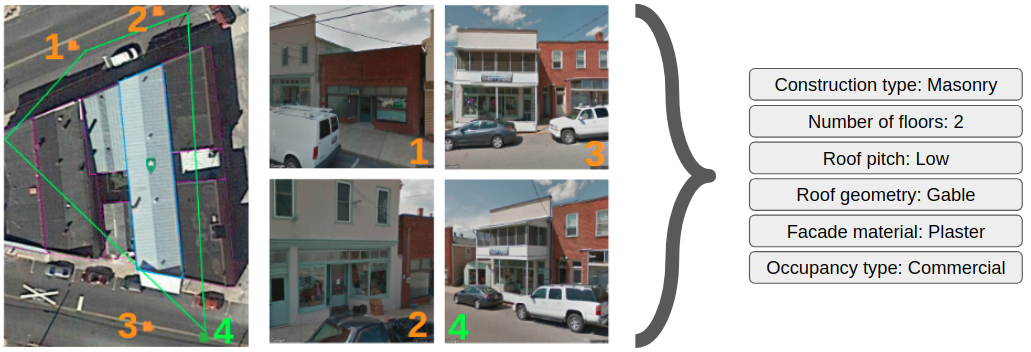}
\caption[An example of a challenging building scene in our dataset.]{An example of a challenging building scene in our dataset. We aim to answer multiple questions about the building, and the answer to each lies in a different input image. The green mark represents the source location of the bottom right image number 4, while the green rays represent its field of view. The building of interest is marked with a green pointer on the top-view image, and it is the white one on the side view. The shading on the top-view image reveals that the roof is not flat, and the geometry is a gable. The front wall occludes most of the roof, but we can still infer its low pitch. The first image is not precisely calibrated and could suggest wrong answers, but combining information from all the sources should prevent it.}
\label{fig:complex}
\end{figure*}

This work aims to develop a deeper understanding of scenes containing buildings based on aerial (top-view) and multiple side view (street-view) images. This problem is both technically challenging and of great practical importance. It allows for automatic pricing of an insurance policy, mechanical claims analysis, risk detection, understanding the environment for self-driving cars, or socioeconomic statistics extraction. The onset of new computer vision techniques and growing large-scale datasets of urban scenes naturally lead to various new scientific challenges related to holistic building understanding, much deeper than simple single-task classification. We investigate a multi-task classification problem of automated building analysis. The goal is to determine the following set of building attributes: construction type, i.e., the material the building is made of, number of floors, roof pitch, roof geometry, facade material, and occupancy type. Those features are crucial for catastrophic risk estimation \cite{stone2018exposure}. The real-world unconstrained environment is often poorly recognizable based on a single image. Rather, it requires more complex analyses of multiple images taken from different angles. Our model uses both the top-view image and numerous street-view photographs of the same building to understand its fine-grained details. 

In this work, we propose a new fusion technique, leveraging both the scene's geometric structure and the high-dimensional features extracted from the top and street-view images. As a result, we obtain a single unified top-view representation of the scene, including information from side views, building outline, and imagery metadata. Based on the street-view photo location, direction, and the field of view, i.e., angle representing the visible range, we construct a projection of the street-view features onto the building walls outlined in the top-view image.

Understanding of physical structures, such as buildings, may require integrating information from all possible input sources. An example of a \emph{building scene} from our dataset is presented in Figure~\ref{fig:complex}. The four camera marks on the top-view image represent four different locations where the street-view images were taken. The green mark and the blue boundary on the top-view image point to the building of interest. The violet boundaries represent other visible buildings, which can be a source of confusion for the model. We can easily see how different images complement each other. The top-view image allows us to see that the roof is not flat as there is a ridge joining the two opposite sides and a skewed shadow. The roof geometry is only visible from the top-view image. The construction type, number of floors, facade material, and occupancy class can be determined only based on the street-view images. 

Some decisions may require more sophisticated reasoning. A possible overlap of the features from different street-view photos might mitigate the errors caused by occlusions and inaccuracies present in a single image. In the example in Figure~\ref{fig:complex}, the first street-view image points towards the center of the building but is occluded by an adjacent building and could lead to classification errors. However, when looking at all four images, it is possible to classify the building's attributes correctly by combining several clues. The side-attached structure made of bricks may correctly suggest that all buildings' construction type along the street is masonry. We are not able to estimate the exact roof pitch by looking exclusively at the street-view images. We can only assume that the roof is flat or has a low slope as it does not stick out from the front. 

To summarize, our contributions are the following:

(1) We propose a new layer called \emph{projection pooling}, which benefits from the building scene geometry and the relationship between images from different views and perspectives. We then integrate it into a deep learning architecture. It results in a new, unified, high-dimensional representation and suits well the classification task. 

(2) We develop a new deep convolutional model, fusing multiple inputs to understand building characteristics better. It includes the projection pooling layer and achieves results that are superior to highly tuned baseline models. These results indicate that one can design substantially more accurate models by incorporating information from multiple images.

(3) We build a new real-world multi-view multi-task dataset of building images, annotations, and attributes. Buildings are heterogeneous in architectural style, size, age, and come from all around the world. It sets a new benchmark of detailed building understanding to spark further research in this area.

\section{Related work}\label{sec:related_work}
In this section, we discuss the ideas which inspired this work. These include (a) the use of street-view images, (b) building modeling, (c) the fusion of street-view and top-view imagery, and (d) building attribute classification.

\textbf{(a)} As street-view imagery is becoming more ubiquitous, it brings new research opportunities. One can assess socioeconomic statistics \cite{gebru2017using,gebru2017fine}, evaluate the safety, beauty and popularity of the neighborhood \cite{convolutional2017investigating,dubey2016deep}, estimate the road safety \cite{song2018farsa}, perform road scene segmentation \cite{cordts2016cityscapes}, and determine precise geolocalization of the car \cite{armagan2017accurate,hirzer2017efficient,armagan2017learning}. Combining house rental ads and street-view imagery allows for performing 3D building reconstruction \cite{chu2016housecraft}. In our work, we show that street-view images can also help with understanding the state of urban structures.

\textbf{(b)} Building facade segmentation is a well-studied problem on 2D images \cite{yang2011hierarchical,tylevcek2013spatial,riemenschneider2012irregular,teboul2011shape,martinovic2013bayesian,kozinski2015mrf,mathias2016atlas,liu2017deepfacade}, through combination of 2D and 3D \cite{gadde2017efficient,riemenschneider2014learning} and directly using 3D data from scanners \cite{serna2016segmentation,li2016extraction}. Many methods rely on assuming symmetry of the building facades \cite{cohen2017symmetry,mitra2013symmetry,musialski2009symmetry,wu2010detecting,zhang2013layered}. Facade datasets are usually collected at a specific location, consist only of a few hundred images, are homogeneous in style and well rectified. Therefore, models can fail when tested on different architectural styles \cite{lotte20183d}. Our work introduces a diverse and large-scale dataset, with buildings coming from all over the world, along with potential noise, and without any assumption about rectification, symmetry or style.

\textbf{(c)} A fusion of street-view and top-view imagery may further improve performance. One can fuse top-view and ground-level imagery for detailed city reconstruction \cite{bodis2016efficient}, use cross-view matching with top-view images to improve street-view geolocalization \cite{workman2015wide,hu2018cvm}, or use cross-view matching with street-view to retrieve latent representation of top-view images \cite{DBLP:journals/corr/abs-1708-03035,cao2018integrating}. In \cite{zhai2017predicting}, the authors transform a semantic top-view scene into a semantic street-view scene. Detecting trees can be achieved through merging both of these sources \cite{DBLP:conf/cvpr/WegnerBHSP16}.
Projecting the street-view latent representation onto an orthographic projection allows for 3D object detection \cite{roddick2018orthographic}. MVSNet \cite{DBLP:conf/eccv/YaoLLFQ18} introduces 'differentiable homography', our layer is a special case of theirs, but we can utilize multiple images in a single training iteration and encode problem constraints directly in the model. Our solution also takes advantage of both input modalities but focuses on detailed building understanding. As opposed to the previous works, our fusion strategy directly utilizes the building geometry to create a 2D latent representation through the projection pooling layer. 

\textbf{(d)} Closely related to our study is the classification of building age, condition, and land use from a single street-view image \cite{zeppelzauer2018automatic,koch2018visual,kang2018building,zhu2019fine}. Urban zone classification was also explored in multi-view settings \cite{srivastava2018fine,hoffmann2019model,srivastava2019understanding}, which we use as our multi-view multi-task baseline models. We are the first to study buildings from all around the world, where multiple recognition tasks are trained together. Most importantly, we propose a new fusion strategy based on building geometry, which boosts our models' classification accuracy.

\section{Dataset}\label{sec:dataset}
One of the goals of this study is to create the first large-scale benchmark of buildings that represent a diverse spectrum of architectural styles, locations, and building attributes. Building sizes vary: from small wooden houses, through churches made of brick, to extensive concrete manufacturing facilities. Thus, models trained on such a dataset should apply to any building in the world. We present the world heatmap of building sites in Appendix A.

Our dataset of top-view images was collected from Google Maps and Bing Maps, while the street-view photos come from Google Street View and Bing StreetSide. The resolution of top-view images is around 30 cm per pixel; see Appendix B for more information on aerial images' resolution. We built a custom tool and annotated the georeferenced top-view images with the following set of classes: buildings, temporary structures, trees. We selected the georeferenced street-view photos based on tree locations and other buildings visible on the top-view image to account for potential occlusions. The images were selected to have the best building visibility and cover all sides of the building. For every building, we have one top-view image and between $1$ and $9$ street-view photos. Knowing the street-view images' locations, we compute the field of view required for every image to capture the entire building. Since these locations can be imprecise, we set the actual field of view to be $20\%$ broader to make sure the whole building is present in the photo. Top and street-view images usually come from different dates, but our annotators verified that both images refer to the same structure. We annotated construction type, number of floors, roof pitch, roof geometry, facade material, and occupancy type for every building of interest. Hired contractors double-checked all the annotations and building characteristics. Additionally, construction types were checked by a professional architect. \\ \\
We classified the building attributes as follows \cite{stone2018exposure}:
\begin{itemize}
\item the construction type describes the predominant material used for the building construction, and it consists of four classes: masonry, metal, reinforced concrete, and wood;
\item the number of floors includes five categories: one, two, three, four and five or more;
\item the roof pitch was divided into four categories: flat ($0^{\circ}$ - $9.5^{\circ}$), low ($9.5^{\circ}$ - $22.5^{\circ}$), medium ($22.5^{\circ}$ - $37^{\circ}$) and steep ($>37^{\circ}$);
\item the roof geometry classes are: flat, gable, hip and shed;
\item facade material was grouped into brick, cement block, concrete, glass, metal, plaster, plastic, stone, and wood;
\item the occupancy classes are the following: agriculture, commercial, industrial, mercantile, public, and residential.
\end{itemize}

Our dataset consists of 6477 training and 3197 testing building scenes, split by stratified sampling. The total number of street-view images is 29350 in the training dataset and 10402 in the testing dataset. Most street-view photos are of size $640\times640$, but around $25\%$ of them are taller. This is because we extended some of the images upwards -- by stitching together multiple photos -- to obtain a better view of taller structures. The size of the top-view image (one per building) depends on the building size.

In summary, our dataset includes the following: top-view images, street-view images, the location, direction and field of view of the street-view images, and the georeferenced footprint of the building.

In Figure \ref{fig:complex} we present a full example, showing all images and metadata available for a single building, along with the correct output classes.

In Appendix C, we present more information about the dataset: an example of each class for each attribute, a sample of full scenes with descriptions, and a comparison with existing datasets from the literature.

\section{Multi-view multi-task approach}\label{sec:multi-view-multi-task}
We start by introducing the basic definitions and then present increasingly strong models for building attribute classification.

We denote the top-view image by $\text{Im}_0$, and the $n$-th street-view input image by $\text{Im}_n$, for $n\in\{1,\ldots,N\}$, where $N$ is the total number of street-view images for the given building. We transform the input images to obtain spatial feature maps $f_0$, $f_1$, ..., $f_N$ using feature extractor networks $\text{CNN}_{\text{TV}}$ (top-view) and $\text{CNN}_{\text{SV}}$ (street-view), with $d$ output channels for street-view features and $d_0$ output channels for top-view features:
\begin{equation} \label{eq:features} 
f_0 = \text{CNN}_{\text{TV}}(\text{Im}_0) \quad f_n = \text{CNN}_{\text{SV}}(\text{Im}_n).
\end{equation}
Feature vectors obtained from these feature maps by mean-pooling along spatial dimensions are denoted as $v_0,\ldots, v_N$. We use six linear heads with softmax activation to generate the class probabilities for each of the tasks. We train the models to minimize the total cross-entropy loss summed over all of the classification tasks. We consider the following baselines.

\begin{figure*}[ht]
\centering
\includegraphics[width=0.8\linewidth]{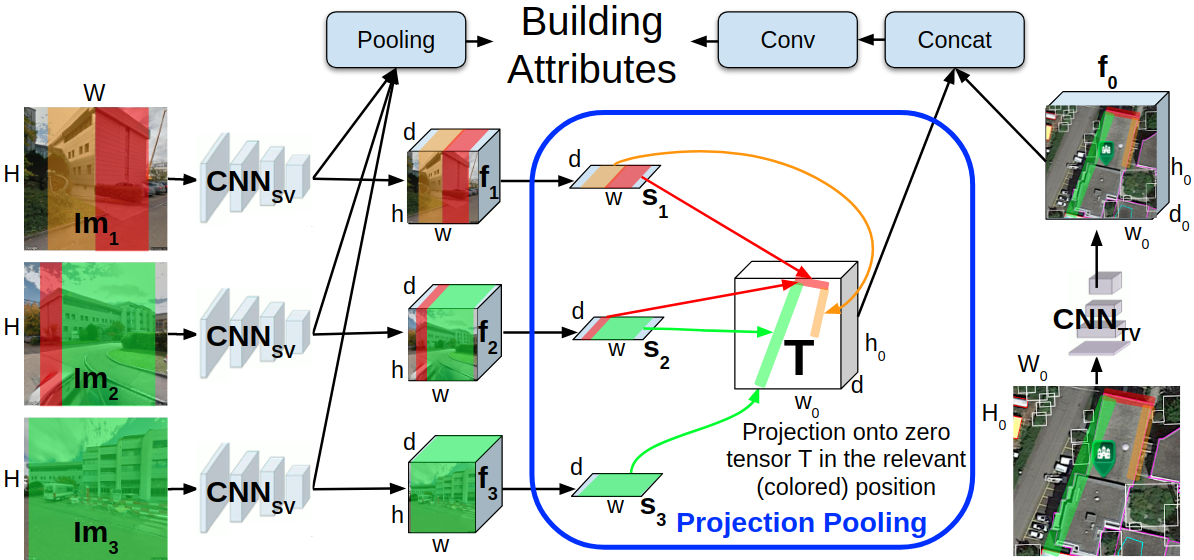} 
\caption[Projection pooling layer.]{Our model with the projection pooling layer. The colors represent different building walls. The red and green wall features come from two different street-view images, while the orange one only from a single photo. The orange segment covers only the visible part of the building wall, as the rest of it is not visible on any street-view image. To overlap them in the top-view projection, we average the street-view features across the height dimension. We project the street-view (side walls) features onto a zero tensor (white). Then, we concatenate the projected features with the top-view feature map to form a unified representation. We further apply convolutions on top of it and concatenate with street-view features to obtain features for the final recognition.}
\label{fig:projection}
\end{figure*}

\textbf{Top-view (TV)} Only the top-view input image is used. The input to the six task-specific classification layers is $v_0$.

\textbf{Street-view (SV)} To fuse the street-view input images, we compute feature vectors $v_1, \ldots, v_N$, and use their average as input to the task-specific classification layers.

\textbf{Street-view + Top-view separately (SV+TV separately)} The different tasks achieve varying results depending on which kinds of images are given as input: street-view or top-view. Based on the results of the baselines that use a single input type (either street-view or top-view), we choose the best baseline model for each task separately. We train the top-view model to classify the roof slope and roof geometry, while the street-view model to classify other attributes. The two networks are completely separate and are trained separately. 

\textbf{Street-view + Top-view (SV+TV)} We use two independent feature extractors: one for street-view and one for top-view images. The vectors $v_1,\ldots, v_N$ are averaged as in the street-view baseline, and then this average is concatenated with $v_0$ to form the input to the task-specific classification layers. This fusion is shown visually in Appendix D. We also consider a modification where the same feature extractor network is used both for the top-view and street-view images (i.e. $\text{CNN}_{\text{TV}}=\text{CNN}_{\text{SV}}$).

\section{Our model}
\label{sec:pooling}
The models described in Section~\ref{sec:multi-view-multi-task} take into account neither the relatively simple cuboid-like geometry of a typical building nor the source positions of the street-view images. These factors, especially the distance between the building and the source of the street-view photo, may impact what is visible on the input image. We design a fusion strategy to leverage the building geometry, street-view photo location, and its direction. It results in a single, unified, top-view representation of the building. We extract the feature maps from all street-view photos as defined in Equation (\ref{eq:features}), and map these features (we call it a \emph{projection} in this work) onto relevant parts of the building polygon, as seen on the top-view image. Those projected features are overlapped and concatenated with the top-view aerial representation to let the model infer the entire scene.

The final representation encompasses the following information: all street-view images, top-view image, building footprint, street-view image positions, directions, and fields of view. 

\subsection{Projection Pooling}\label{sec:pooling_layer}
We describe the construction of the unified building representation following the explanatory Figure \ref{fig:projection}. For each street-view image $\text{Im}_i$ of size $[H_i, W_i, 3]$, we extract the feature map $f_i$ of shape $[h_i, w_i, d]$, and average out the vertical dimension, while keeping the horizontal dimension intact. It results in a feature stripe $s_i$ of shape $[w_i, d]$, which will be projected onto the building outline. Since all the street-view images in our dataset are of the same width, we can replace the $w_i$'s by one value $w$, so all the stripes $s_i$ are in fact of shape $[w, d]$. We initialize a zero tensor $T$ of shape $[h_0, w_0, d]$, where $h_0$ and $w_0$ are the height and width of the feature map $f_0$. $T$ represents a top-view 2D-grid of $d$-dimensional neurons. Neurons in $T$ are projected from each stripe $s_i$. By concatenating $T$ of shape $[h_0, w_0, d]$ with the top-view feature map $f_0$ of shape $[h_0, w_0, d_0]$, we construct the final unified representation.

In this paragraph, we explain the details of how we project a stripe $s_i$ onto the tensor $T$. Using the building outline, source location, and direction of street-view image $\text{Im}_i$, we compute the visible parts of the building polygon (represented by colored segments on tensor $T$ in Figure~\ref{fig:projection}). An efficient way of computing them is described in Appendix E. These segments correspond to parts of the edges of the building polygon visible in $\text{Im}_0$. They can be discretized and approximated by a set $P_i$ of $d$-dimensional neurons located in $[h_0, w_0]$ 2D-grid of $T$. For each neuron $p=(x_p, y_p)\in P_i$, we compute the angle at which it can be seen from the location of $\text{Im}_i$. This forms its cone of visibility, as presented in Figure \ref{fig:sampling}. This cone lets us compute the visible part of the building corresponding to $p$ in the image $\text{Im}_i$. This translates to an angle range $(a_l, a_r)$ along the width dimension of $s_i$. We project the features from this range in $s_i$ onto the $(x_p, y_p)$ position in $T$. In Subsection~\ref{sec:sampling}, we explore various ways of sampling the features from a single stripe $s_i$. Some of the neurons in $T$ might not be visible on any of the street-view images, or do not belong to the building polygon - the corresponding feature vectors in $T$ will be zero vectors. If a single neuron in $T$ gets features projected from multiple stripes, these feature vectors are max-pooled.

This description can be formulated as follows:
\begin{equation*}
s_i[w^{\prime},d^{\prime}] = \frac{\sum_{h^{\prime} \in [0 \ldots h_i-1]} f_i[h^{\prime},w^{\prime},d^{\prime}]}{h_i}
\end{equation*}
\begin{equation*}
T[h^{\prime}, w^{\prime}] = \max\limits_{i \in [1 \ldots N]} \mathrm{Pool}(h^{\prime}, w^{\prime}, prt, s_i, src_i, dir_i, FoV_i)
\end{equation*}

\begin{algorithm}
\caption{Pool$(h^{\prime}, w^{\prime}, prt, s_i, src_i, dir_i, FoV_i)$}
\begin{algorithmic}
    \STATE $p$ = $(h^{\prime}, w^{\prime})$
    \IF {$p$ not in ConeOfVisibility($src_i, dir_i, FoV_i$)}
        \RETURN $0$
    \ENDIF
    \IF {$p$ not in Boundary($prt$)}
        \RETURN $0$
    \ENDIF
    \IF {$p$ in OccludedByOtherWall($prt, src_i, dir_i, FoV_i$)}
        \RETURN $0$
    \ENDIF
    \STATE $a_l, a_c, a_r$ = PixelToStripeRange($p, prt, src_i, dir_i, FoV_i$)
    \RETURN StripeSample($a_l, a_c, a_r, w, s_i, FoV_i$) 
\end{algorithmic}
\end{algorithm}

Where $h^{\prime}, w^{\prime}, d^{\prime}$ are indexes of height, width and depth dimensions, $prt$ is the building footprint, $N$ is number of street-view images, $src_i$ is source location of i-th street-view image, $dir_i$ is direction of i-th street-view image and $FoV_i$ is its field of view.

\subsection{Stripe sampling (SS)}
\label{sec:sampling}
\begin{figure}[t]
\centering
\includegraphics[width=\linewidth]{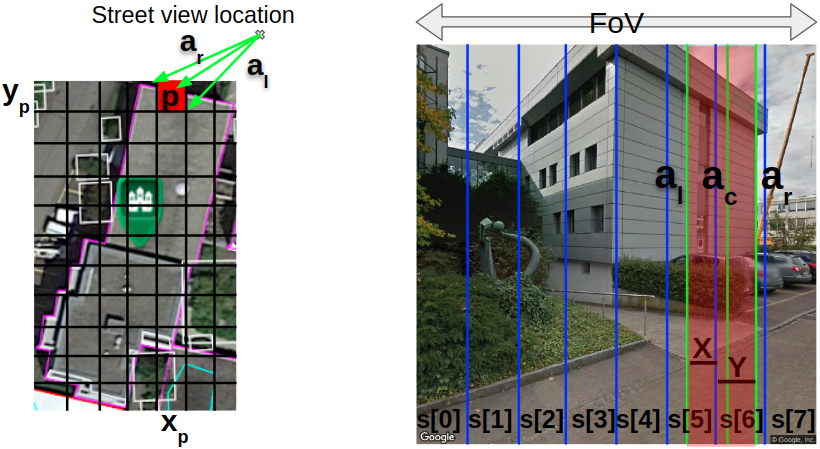} 
\caption[Stripe sampling.]{Different strategies for sampling stripe features in the projection pooling layer. The red box represents a neuron in 2D-grid $T$. The green rays towards the neuron center and its sides are also marked as green vertical lines on the street-view image. $s$ is defined as the average along the height dimension of street-view features. In this example, we assume $w=8$. The 'nearest' strategy takes the nearest position in the stripe i.e. $s[6]$. The 'sum' strategy takes the weighted sum between pixel boundaries $X s[5] + Y s[6]$. The 'avg' normalizes the 'sum' strategy i.e. $(X s[5] + Y s[6])/(X+Y)$. }
\label{fig:sampling}
\end{figure}

A single neuron in the top-view feature map can correspond to a relatively large area in the original input image. The angle of visibility for a neuron on a street-view image can cover multiple pixels in $s_i$, especially for the close-by photos (see the red area in Figure \ref{fig:sampling}). For example, a $1\times 1$ pixel in the output feature map of ResNet-50 \cite{DBLP:journals/corr/HeZRS15} with a total stride of 32 corresponds to an area of $32\times 32$ pixels in the input image. The details on how to calculate the neuron value can be crucial for the overall performance of the model \cite{DBLP:journals/corr/HeGDG17}. We define three different strategies for deriving the neuron value from the $s_i$ stripes. While we describe these strategies below, we also presented them visually in Figure \ref{fig:sampling}. In all strategies, $FoV$ represents the field of view angle for the entire street-view image, $w$ is the width of the associated stripe $s$, and $a_x^{\prime} = w\frac{a_x}{FoV}$ for $x \in \{l, r, c\}$.

\begin{figure*}[t]
\centering
\includegraphics[width=0.7\linewidth]{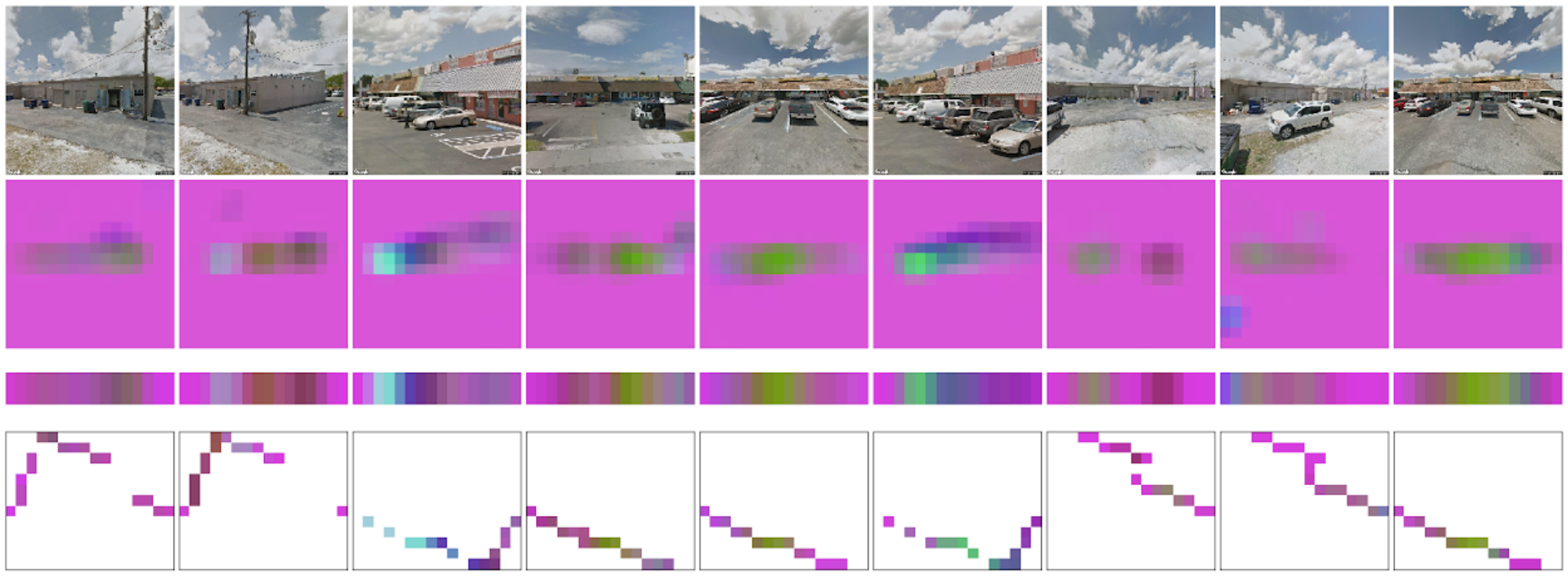} 
\includegraphics[width=0.1575\linewidth]{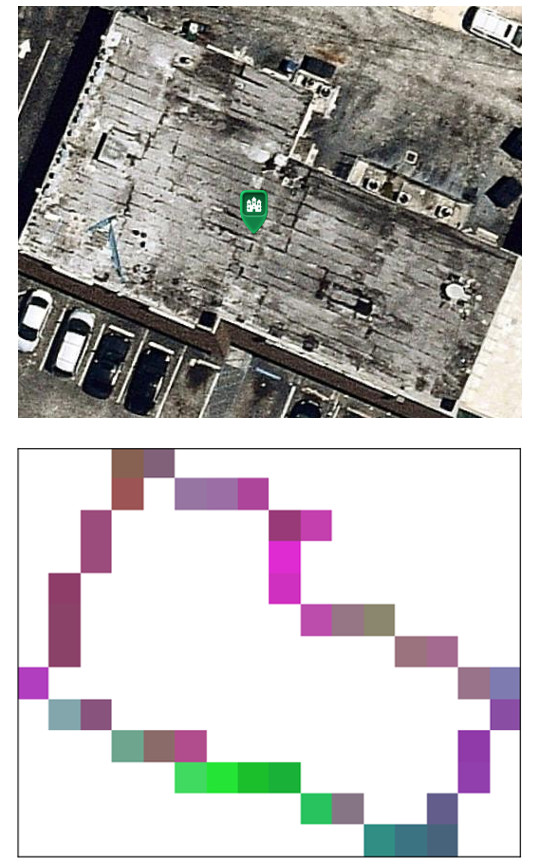}  
\caption[Visualization of projection pooling features.]{Left: We show street-view photos corresponding to a single building (top row), visualization of the extracted feature maps $f_i$ before averaging out the height dimension (second row), and after (third row). In the last row, we show features contributed to specific neurons of the unified representation. Right: Visualization of the unified representation pooled from all the images. The neurons shown as white did not get any features projected onto them, i.e., their values are $0$. For a larger version of this figure, see Appendix F.}
\label{fig:projections}
\end{figure*}

\textbf{Nearest (N)} uses the angle, measured clockwise, between the beginning of the field of view of given street-view image and the center of the neuron. We denote this angle by $a_c$. In this strategy, we use a single feature vector from $s$ to calculate the value of neuron $p$:
\begin{equation}  \label{eq:nearest} 
p_{\text{nearest}} = s(\lfloor a_c^{\prime} \rfloor).
\end{equation}
In other words, we draw a ray from the location of the street-view photo towards the neuron's center, calculate the position $p\in\{0, \ldots, w-1\}$ of its intersection with the stripe $s$, and take the $d$-dimensional feature vector at this position (recall that $s$ is of shape $[w, d]$).

\textbf{Sum (S)} uses two rays pointing towards the pixel's ends instead of one ray pointing towards the pixel's center. Denote the angles of these rays by $a_l$ and $a_r$ for the left and right one, respectively. Then, the value of the projected feature vector is defined below, where $\{x\} = x - \lfloor x \rfloor$.
\begin{equation}  \label{eq:sum} 
p_{\text{sum}} = s(\lfloor a_l^{\prime} \rfloor) \cdot \{-a_l^{\prime}\} +
           s(\lfloor a_r^{\prime}\rfloor) \cdot \{a_r^{\prime} \} + \sum_{i=\lceil a_l^{\prime} \rceil}^{\lfloor a_r^{\prime} \rfloor - 1} s(i)
\end{equation}

\textbf{Average (A)} follows a similar strategy to Sum -- i.e. it takes into account feature vectors from multiple positions along the feature stripe $s$. However, a potential problem with the Sum strategy is that the magnitude of values in the resulting vector may vary significantly with the distance to the building wall: more feature vectors are summed if the building wall is in proximity. Therefore, we propose a variant of the Sum strategy, where the output is additionally normalized. We choose to simply divide the resulting feature vector by the width of the part of $s$ corresponding to the neuron $p$:
\begin{equation} \label{eq:avg} 
p_{\text{average}} = \frac{p_{\text{sum}}}{a_r^{\prime} - a_l^{\prime}}
\end{equation}

\subsection{Visualizing projected features}
We visualize the features extracted from the street-view photos and how they are projected to create the final unified building representation. We obtain features from the test set pictures corresponding to multiple buildings and project them to three dimensions using principal component analysis, resulting in RGB coordinates. It allows us to visualize a single feature vector as a color.

We visualize the street-view feature maps both before and after their height dimension is averaged out. Then, we look at the feature vectors projected onto the building polygon from each street-view photo separately. We show this in Figure~\ref{fig:projections}. Note that the south wall looks substantially different from the others for the selected building, and this difference can be seen in the visualizations. Moreover, feature maps in positions where no structure is visible get projected to the same uniform color (pink), showing that the feature extractor ignores irrelevant information. Finally, we examine the unified representation vectors, i.e., after max-pooling the features projected from different views. Again, we see that the side of the building providing different visual information has a substantially different projection.

\begin{table*}[t]
\centering
\begin{tabular}{lccccccc}
\toprule
&  \multicolumn{7}{c}{Testing Accuracy} \\
\midrule
Model &  Constr & \#Floors & Roof Pitch & Roof Geom & Facade Material & Occup &  Avg \\
\midrule
TV - only top-view      & 72.2 & 57.2 & 79.2 & 90.9 & 51.1 & 70.0 & \textbf{70.10} \\ 
SV - only street-view    & 73.4 & 72.2 & 76.7 & 87.0 & 61.5 & 74.2 & \textbf{74.14} \\
SV+TV, $\text{CNN}_{\text{TV}}=\text{CNN}_{\text{SV}}$    & 73.1 & 72.0 & 79.7 & 91.1 & 58.5 & 72.5 & \textbf{74.49} \\
SV+TV           &    73.0 & 72.7 & 79.4 & 91.4 & 57.9 & 73.2 & \textbf{74.59} \\
SV+TV, separately     & 73.3 & 72.5 & 77.7 & 89.8 & 60.3 & 74.2 & \textbf{74.64} \\
Projection pooling  & 76.0 & 75.6 & 81.3 & 91.9 & 62.4 & 76.6 & \textbf{77.28} \\
\bottomrule
\end{tabular}
\caption[The classification comparison for different methods.]{The classification accuracy results of our method and the baselines.}
\label{tab:results}
\end{table*}

\section{Experiments and results}
\subsection{Experimental setup}
We benchmark our approach on the newly collected dataset of building attributes, as defined in Section \ref{sec:dataset}. Street-view images are resized to size $500\times 500$. The top-view image is cropped to the bounding box containing the building of interest and resized to keep the aspect ratio so that the length of the longer side is 500. As the feature extractor network, we use the ResNet-50 model pre-trained on the ImageNet dataset, as available in PyTorch \cite{paszke2017automatic}. We discard the final fully connected output layer. We freeze the parameters of the stem and the first block in the pre-trained ResNet-50 feature extractors and do not fine-tune the batch normalization layers. We train the network using stochastic gradient descent with momentum, on a single GPU (see Appendix G for details), and with a batch size of one building. The effective number of images in a single batch is equal to 1 for the top-view branch, while for the street-view branch, it varies from 1 to 9. We use the learning rate of 0.0001 for ten epochs and then 0.00001 for one more epoch. We set the momentum to 0.9. We apply L2 regularization with a weight decay of 0.001 and augment the dataset with random color jittering. We compare the models using average classification accuracy over the six tasks. For details about hyperparameter tuning, see Appendix H.

\textbf{Image dropout (ID)} Inspired by dropout \cite{srivastava2014dropout}, we regularize the training by randomly dropping entire street-view images, which we call \textit{image dropout}. During training, every street-view image in the batch of a given building is omitted with probability $p$. We make sure that there is always at least one street-view image in a batch. We do not rescale the values of the features as done in the original dropout. With image dropout, we force the model to use a different set of images in each iteration. We do not apply image dropout at test time, allowing the model to use all of the street-view images available for the given building.

\textbf{Projection thickness (TH)} When rounding up the top-view building polygon to a set of neurons in $T$, we obtain a thin building outline, where each side is discretized as a one-pixel wide line. We consider using an outline wider than one. Projection thickness of $k$ means that the building outline is discretized to line segments, which are $k$ neurons wide. We assume that pixels belonging to the same side do not occlude each other. Widening the polygon makes it easier for the convolutional layer applied on top of the unified representation to capture features from adjacent walls.

\textbf{Cutout (CU)} To further regularize the training over multiple potentially redundant street-view images, we apply Cutout \cite{devries2017improved}. For every street-view image, with probability $q$, we blackout $40\%$ of pixels, by covering the image with a randomly placed black rectangle. Using this approach, we force the model to rely on multiple parts of an image when performing classification. 

\textbf{Splitting the street-view images (SS)} \label{sec:ss} To perform projection pooling, we average the street-view feature maps $f_i$ along the height dimension, which gives a rough feature stripe $s_i$. We also investigate a different strategy, which allows the model to take into account the differences between lower, middle, and upper parts of the street-view images. We split the feature representation into $k$ different tensors along the height dimension, and concatenate them along the depth dimension. In other words, we shift the height dimension into the depth dimension. From a feature map $f_i$ of size $[h, w, d]$, we obtain one of size $[\frac{h}{k}, w, d\cdot k]$ and apply mean-pooling along height dimension with projection pooling. 

\subsection{Results}
In this subsection, we discuss the experimental results.

In Table~\ref{tab:results}, we present a comparison between competitive methods and the new proposed model with the projection pooling layer. Using only the top-view images gives worse results than using only the street-view photos. It is still better than one would expect, given that most parts of the building are not visible on the top-view image. Unsurprisingly, using the street-view rather than the top-view gives the most substantial gains when counting the number of floors (+15\%) and predicting facade material (+10.4\%). On the other hand, top-view images are more informative when it comes to roof geometry (+3.9\%) and roof pitch (+2.5\%).

Combining both top-view and street-view images gives better results than using a single source, but only by 0.35\% than when using street-view alone. Separate networks for the top and street-view images yield a small improvement. We obtain the best baseline results when training the top-view and street-view networks independently on different tasks. 

Incorporating the projection pooling layer results in about 3\% accuracy improvement over the best baseline.

In Appendix I, we present the results of an ablation study, which examines the impact of all components of our final model. First, we train a vanilla projection pooling model with different stripe sampling strategies, and find that using the averaging strategy gives the best results (75.93\%). Adding image dropout with probability $p=50\%$ further regularizes the model, and increases average accuracy (+0.23\%). We then tested values higher than $1$ for projection thickness, and found that using $3$ improves performance (+0.38\%), while increasing beyond this number did not bring further gains. Applying cutout with probability $q=50\%$ improves our model even further (+0.33\%), which shows the importance of using a wide variety of regularization techniques. The final improvement (+0.41\%) comes from splitting the street-view feature maps into three tensors, and concatenating them along the depth dimension, as described in Subsection~\ref{sec:ss}. In this way, lower, middle, and upper parts of the building are separately projected, which is especially helpful for discrimination of the number of floors.

\textbf{Impact of multiple images} We investigate the impact of varying the number of street-view images per building. For $k \in \{1, 2, 3, 4\}$, we test our best model with a restriction to use at most $k$ street-view images.

The results of this comparison are shown in Appendix J. We see a substantial (+1.36\%) gain from using more than one street-view image, suggesting that photos from multiple angles are often necessary for correct classification. On the other hand, the benefits of adding more street-view images quickly plateau, as there is a considerable overlap of information provided by the different street-view photos.

\section{Conclusions}
Our study presents a novel solution to a practical problem of building understanding. For the first time, this problem is approached using the building geometry inside a deep neural architecture to create a unified high-dimensional representation. We propose a new way to integrate the features from multiple views called \textit{projection pooling}. It is a general method for creating a unified representation of 3D objects from orthogonal projections and is particularly well suited for building analysis. In the future, it can be tested against analogical setups, such as mammogram analysis \cite{morrell2018large}. We propose a model for building feature recognition, which incorporates the projection pooling layer, and its results are superior to highly tuned baseline models.

We build a new dataset with fine-grained building attributes and analyze techniques for integrating information from multiple views. The dataset establishes a demanding benchmark for state-of-the-art deep learning methods. It requires reasoning about numerous images at once to give accurate results. We plan to expand the dataset to detect objects, such as doors and windows.

\bibliography{references}

\begin{thebibliography}{51}
\providecommand{\natexlab}[1]{#1}
\providecommand{\url}[1]{\texttt{#1}}
\providecommand{\urlprefix}{URL }
\expandafter\ifx\csname urlstyle\endcsname\relax
  \providecommand{\doi}[1]{doi:\discretionary{}{}{}#1}\else
  \providecommand{\doi}{doi:\discretionary{}{}{}\begingroup
  \urlstyle{rm}\Url}\fi

\bibitem[{Andersson, Birck, and Araujo(2017)}]{convolutional2017investigating}
Andersson, V.~O.; Birck, M.~A.; and Araujo, R.~M. 2017.
\newblock Investigating crime rate prediction using street-level images and
  siamese convolutional neural networks.
\newblock In \emph{Latin American Workshop on Computational Neuroscience}.

\bibitem[{Armagan et~al.(2017{\natexlab{a}})Armagan, Hirzer, Roth, and
  Lepetit}]{armagan2017accurate}
Armagan, A.; Hirzer, M.; Roth, P.~M.; and Lepetit, V. 2017{\natexlab{a}}.
\newblock Accurate camera registration in urban environments using high-level
  feature matching.
\newblock In \emph{Proceedings of the British Machine Vision Conference}.

\bibitem[{Armagan et~al.(2017{\natexlab{b}})Armagan, Hirzer, Roth, and
  Lepetit}]{armagan2017learning}
Armagan, A.; Hirzer, M.; Roth, P.~M.; and Lepetit, V. 2017{\natexlab{b}}.
\newblock Learning to align semantic segmentation and 2.5 d maps for
  geolocalization.
\newblock In \emph{Proceedings of the IEEE Conference on Computer Vision and
  Pattern Recognition}, 3425--3432.

\bibitem[{B{\'o}dis-Szomor{\'u}, Riemenschneider, and
  Van~Gool(2016)}]{bodis2016efficient}
B{\'o}dis-Szomor{\'u}, A.; Riemenschneider, H.; and Van~Gool, L. 2016.
\newblock Efficient volumetric fusion of airborne and street-side data for
  urban reconstruction.
\newblock In \emph{International Conference on Pattern Recognition},
  3204--3209. IEEE.

\bibitem[{Cao et~al.(2018)Cao, Zhu, Tu, Li, Cao, Liu, Zhang, and
  Qiu}]{cao2018integrating}
Cao, R.; Zhu, J.; Tu, W.; Li, Q.; Cao, J.; Liu, B.; Zhang, Q.; and Qiu, G.
  2018.
\newblock Integrating Aerial and Street View Images for Urban Land Use
  Classification.
\newblock \emph{Remote Sensing} .

\bibitem[{Chu et~al.(2016)Chu, Wang, Urtasun, and Fidler}]{chu2016housecraft}
Chu, H.; Wang, S.; Urtasun, R.; and Fidler, S. 2016.
\newblock Housecraft: Building houses from rental Ads and street views.
\newblock In \emph{Proceedings of the European Conference on Computer Vision},
  500--516. Springer.

\bibitem[{Cohen et~al.(2017)Cohen, Oswald, Liu, and
  Pollefeys}]{cohen2017symmetry}
Cohen, A.; Oswald, M.~R.; Liu, Y.; and Pollefeys, M. 2017.
\newblock Symmetry-Aware fa{\c{c}}ade parsing with occlusions.
\newblock In \emph{2017 International Conference on 3D Vision (3DV)}.

\bibitem[{Cordts et~al.(2016)Cordts, Omran, Ramos, Rehfeld, Enzweiler,
  Benenson, Franke, Roth, and Schiele}]{cordts2016cityscapes}
Cordts, M.; Omran, M.; Ramos, S.; Rehfeld, T.; Enzweiler, M.; Benenson, R.;
  Franke, U.; Roth, S.; and Schiele, B. 2016.
\newblock The cityscapes dataset for semantic urban scene understanding.
\newblock In \emph{Proceedings of the IEEE Conference on Computer Vision and
  Pattern Recognition}, 3213--3223.

\bibitem[{DeVries and Taylor(2017)}]{devries2017improved}
DeVries, T.; and Taylor, G.~W. 2017.
\newblock Improved regularization of convolutional neural networks with cutout.
\newblock \emph{arXiv preprint arXiv:1708.04552} .

\bibitem[{Dubey et~al.(2016)Dubey, Naik, Parikh, Raskar, and
  Hidalgo}]{dubey2016deep}
Dubey, A.; Naik, N.; Parikh, D.; Raskar, R.; and Hidalgo, C.~A. 2016.
\newblock Deep learning the city: Quantifying urban perception at a global
  scale.
\newblock In \emph{Proceedings of the European Conference on Computer Vision},
  196--212. Springer.

\bibitem[{Gadde et~al.(2017)Gadde, Jampani, Marlet, and
  Gehler}]{gadde2017efficient}
Gadde, R.; Jampani, V.; Marlet, R.; and Gehler, P.~V. 2017.
\newblock Efficient 2D and 3D facade segmentation using auto-context.
\newblock \emph{IEEE Transactions on Pattern Analysis and Machine Intelligence
  (PAMI)} 40(5): 1273--1280.

\bibitem[{Gebru et~al.(2017{\natexlab{a}})Gebru, Krause, Wang, Chen, Deng,
  Aiden, and Fei-Fei}]{gebru2017using}
Gebru, T.; Krause, J.; Wang, Y.; Chen, D.; Deng, J.; Aiden, E.~L.; and Fei-Fei,
  L. 2017{\natexlab{a}}.
\newblock Using deep learning and Google Street View to estimate the
  demographic makeup of neighborhoods across the United States.
\newblock \emph{Proceedings of the National Academy of Sciences} 114(50):
  13108--13113.

\bibitem[{Gebru et~al.(2017{\natexlab{b}})Gebru, Krause, Wang, Chen, Deng, and
  Fei-Fei}]{gebru2017fine}
Gebru, T.; Krause, J.; Wang, Y.; Chen, D.; Deng, J.; and Fei-Fei, L.
  2017{\natexlab{b}}.
\newblock Fine-grained car detection for visual census estimation.
\newblock In \emph{Thirty-First AAAI Conference on Artificial Intelligence}.

\bibitem[{He et~al.(2017)He, Gkioxari, Doll{\'{a}}r, and
  Girshick}]{DBLP:journals/corr/HeGDG17}
He, K.; Gkioxari, G.; Doll{\'{a}}r, P.; and Girshick, R.~B. 2017.
\newblock Mask {R-CNN}.
\newblock In \emph{Proceedings of the IEEE International Conference on Computer
  Vision}, 2961--2969.

\bibitem[{He et~al.(2016)He, Zhang, Ren, and Sun}]{DBLP:journals/corr/HeZRS15}
He, K.; Zhang, X.; Ren, S.; and Sun, J. 2016.
\newblock Deep Residual Learning for Image Recognition.
\newblock In \emph{Proceedings of the IEEE Conference on Computer Vision and
  Pattern Recognition}.

\bibitem[{Hirzer et~al.(2017)Hirzer, Arth, Roth, and
  Lepetit}]{hirzer2017efficient}
Hirzer, M.; Arth, C.; Roth, P.~M.; and Lepetit, V. 2017.
\newblock Efficient 3D Tracking in Urban Environments with Semantic
  Segmentation.
\newblock In \emph{Proceedings of the British Machine Vision Conference}.

\bibitem[{Hoffmann et~al.(2019)Hoffmann, Wang, Werner, Kang, and
  Zhu}]{hoffmann2019model}
Hoffmann, E.~J.; Wang, Y.; Werner, M.; Kang, J.; and Zhu, X.~X. 2019.
\newblock Model Fusion for Building Type Classification from Aerial and Street
  View Images.
\newblock \emph{Remote Sensing} .

\bibitem[{Hu et~al.(2018)Hu, Feng, Nguyen, and Hee~Lee}]{hu2018cvm}
Hu, S.; Feng, M.; Nguyen, R.~M.; and Hee~Lee, G. 2018.
\newblock CVM-Net: Cross-View Matching Network for Image-Based Ground-to-Aerial
  Geo-Localization.
\newblock In \emph{Proceedings of the IEEE Conference on Computer Vision and
  Pattern Recognition}, 7258--7267.

\bibitem[{Kang et~al.(2018)Kang, K{\"o}rner, Wang, Taubenb{\"o}ck, and
  Zhu}]{kang2018building}
Kang, J.; K{\"o}rner, M.; Wang, Y.; Taubenb{\"o}ck, H.; and Zhu, X.~X. 2018.
\newblock Building instance classification using street view images.
\newblock \emph{ISPRS Journal of Photogrammetry and Remote Sensing} .

\bibitem[{Koch et~al.(2018)Koch, Despotovic, Sakeena, D{\"o}ller, and
  Zeppelzauer}]{koch2018visual}
Koch, D.; Despotovic, M.; Sakeena, M.; D{\"o}ller, M.; and Zeppelzauer, M.
  2018.
\newblock Visual estimation of building condition with patch-level convnets.
\newblock In \emph{Proceedings of the 2018 ACM Workshop on Multimedia for Real
  Estate Tech}, 12--17. ACM.

\bibitem[{Kozinski et~al.(2015)Kozinski, Gadde, Zagoruyko, Obozinski, and
  Marlet}]{kozinski2015mrf}
Kozinski, M.; Gadde, R.; Zagoruyko, S.; Obozinski, G.; and Marlet, R. 2015.
\newblock A MRF shape prior for facade parsing with occlusions.
\newblock In \emph{Proceedings of the IEEE Conference on Computer Vision and
  Pattern Recognition}, 2820--2828.

\bibitem[{Li et~al.(2016)Li, Hu, Wu, Liu, and Wu}]{li2016extraction}
Li, Y.; Hu, Q.; Wu, M.; Liu, J.; and Wu, X. 2016.
\newblock Extraction and simplification of building fa{\c{c}}ade pieces from
  mobile laser scanner point clouds for 3d street view services.
\newblock \emph{ISPRS International Journal of Geo-Information} 5(12): 231.

\bibitem[{Liu et~al.(2017)Liu, Zhang, Zhu, and Hoi}]{liu2017deepfacade}
Liu, H.; Zhang, J.; Zhu, J.; and Hoi, S. C.~H. 2017.
\newblock DeepFacade: {A} Deep Learning Approach to Facade Parsing.
\newblock In \emph{Proceedings of the Twenty-Sixth International Joint
  Conference on Artificial Intelligence, {IJCAI} 2017, Melbourne, Australia,
  August 19-25, 2017}.

\bibitem[{Lotte et~al.(2018)Lotte, Haala, Karpina, Arag{\~a}o, Shimabukuro
  et~al.}]{lotte20183d}
Lotte, R.; Haala, N.; Karpina, M.; Arag{\~a}o, L.; Shimabukuro, Y.; et~al.
  2018.
\newblock 3D Fa{\c{c}}ade Labeling over Complex Scenarios: A Case Study Using
  Convolutional Neural Network and Structure-From-Motion.
\newblock \emph{Remote Sensing} .

\bibitem[{Martinovic and Van~Gool(2013)}]{martinovic2013bayesian}
Martinovic, A.; and Van~Gool, L. 2013.
\newblock Bayesian grammar learning for inverse procedural modeling.
\newblock In \emph{Proceedings of the IEEE Conference on Computer Vision and
  Pattern Recognition}, 201--208.

\bibitem[{Mathias, Martinovi{\'c}, and Van~Gool(2016)}]{mathias2016atlas}
Mathias, M.; Martinovi{\'c}, A.; and Van~Gool, L. 2016.
\newblock ATLAS: A three-layered approach to facade parsing.
\newblock \emph{International Journal of Computer Vision} 118(1): 22--48.

\bibitem[{Mitra et~al.(2012)Mitra, Pauly, Wand, and Ceylan}]{mitra2013symmetry}
Mitra, N.~J.; Pauly, M.; Wand, M.; and Ceylan, D. 2012.
\newblock Symmetry in 3D Geometry: Extraction and Applications.
\newblock In \emph{Eurographics}.

\bibitem[{Morrell et~al.(2018)Morrell, Wojna, Khoo, Ourselin, and
  Iglesias}]{morrell2018large}
Morrell, S.; Wojna, Z.; Khoo, C.~S.; Ourselin, S.; and Iglesias, J.~E. 2018.
\newblock Large-Scale Mammography CAD with Deformable Conv-Nets.
\newblock In \emph{Image Analysis for Moving Organ, Breast, and Thoracic
  Images}, 64--72. Springer.

\bibitem[{Musialski et~al.(2009)Musialski, Wonka, Recheis, Maierhofer, and
  Purgathofer}]{musialski2009symmetry}
Musialski, P.; Wonka, P.; Recheis, M.; Maierhofer, S.; and Purgathofer, W.
  2009.
\newblock Symmetry-Based Fa{\c{c}}ade Repair.
\newblock In \emph{Proceedings of the Vision, Modeling, and Visualization
  Workshop 2009, November 16-18, 2009, Braunschweig, Germany}, 3--10.

\bibitem[{Paszke et~al.(2017)Paszke, Gross, Chintala, Chanan, Yang, DeVito,
  Lin, Desmaison, Antiga, and Lerer}]{paszke2017automatic}
Paszke, A.; Gross, S.; Chintala, S.; Chanan, G.; Yang, E.; DeVito, Z.; Lin, Z.;
  Desmaison, A.; Antiga, L.; and Lerer, A. 2017.
\newblock Automatic differentiation in PyTorch.
\newblock In \emph{NIPS-W}.

\bibitem[{Riemenschneider et~al.(2014)Riemenschneider, B{\'o}dis-Szomor{\'u},
  Weissenberg, and Van~Gool}]{riemenschneider2014learning}
Riemenschneider, H.; B{\'o}dis-Szomor{\'u}, A.; Weissenberg, J.; and Van~Gool,
  L. 2014.
\newblock Learning where to classify in multi-view semantic segmentation.
\newblock In \emph{Proceedings of the European Conference on Computer Vision},
  516--532. Springer.

\bibitem[{Riemenschneider et~al.(2012)Riemenschneider, Krispel, Thaller,
  Donoser, Havemann, Fellner, and Bischof}]{riemenschneider2012irregular}
Riemenschneider, H.; Krispel, U.; Thaller, W.; Donoser, M.; Havemann, S.;
  Fellner, D.; and Bischof, H. 2012.
\newblock Irregular lattices for complex shape grammar facade parsing.
\newblock In \emph{Proceedings of the IEEE Conference on Computer Vision and
  Pattern Recognition}, 1640--1647. IEEE.

\bibitem[{Roddick, Kendall, and Cipolla(2019)}]{roddick2018orthographic}
Roddick, T.; Kendall, A.; and Cipolla, R. 2019.
\newblock Orthographic Feature Transform for Monocular 3D Object Detection.
\newblock In \emph{30th British Machine Vision Conference 2019, {BMVC} 2019,
  Cardiff, UK, September 9-12, 2019}. {BMVA} Press.

\bibitem[{Serna, Marcotegui, and Hern{\'a}ndez(2016)}]{serna2016segmentation}
Serna, A.; Marcotegui, B.; and Hern{\'a}ndez, J. 2016.
\newblock Segmentation of fa{\c{c}}ades from urban 3D point clouds using
  geometrical and morphological attribute-based operators.
\newblock \emph{ISPRS International Journal of Geo-Information} 5(1): 6.

\bibitem[{Song et~al.(2018)Song, Workman, Hadzic, Zhang, Green, Chen,
  Souleyrette, and Jacobs}]{song2018farsa}
Song, W.; Workman, S.; Hadzic, A.; Zhang, X.; Green, E.; Chen, M.; Souleyrette,
  R.; and Jacobs, N. 2018.
\newblock FARSA: Fully Automated Roadway Safety Assessment.
\newblock In \emph{Winter Conference on Applications of Computer Vision
  (WACV)}, 521--529. IEEE.

\bibitem[{Srivastava et~al.(2014)Srivastava, Hinton, Krizhevsky, Sutskever, and
  Salakhutdinov}]{srivastava2014dropout}
Srivastava, N.; Hinton, G.; Krizhevsky, A.; Sutskever, I.; and Salakhutdinov,
  R. 2014.
\newblock Dropout: a simple way to prevent neural networks from overfitting.
\newblock \emph{The Journal of Machine Learning Research} 15(1): 1929--1958.

\bibitem[{Srivastava et~al.(2018)Srivastava, Vargas~Mu{\~n}oz, Lobry, and
  Tuia}]{srivastava2018fine}
Srivastava, S.; Vargas~Mu{\~n}oz, J.~E.; Lobry, S.; and Tuia, D. 2018.
\newblock Fine-grained landuse characterization using ground-based pictures: a
  deep learning solution based on globally available data.
\newblock \emph{International Journal of Geographical Information Science}
  1--20.

\bibitem[{Srivastava, Vargas-Mu{\~n}oz, and
  Tuia(2019)}]{srivastava2019understanding}
Srivastava, S.; Vargas-Mu{\~n}oz, J.~E.; and Tuia, D. 2019.
\newblock Understanding urban landuse from the above and ground perspectives: A
  deep learning, multimodal solution.
\newblock \emph{Remote Sensing of Environment} .

\bibitem[{Stone(2018)}]{stone2018exposure}
Stone, H. 2018.
\newblock \emph{Exposure and vulnerability for seismic risk evaluations}.
\newblock Ph.D. thesis, UCL (University College London).

\bibitem[{Teboul et~al.(2011)Teboul, Kokkinos, Simon, Koutsourakis, and
  Paragios}]{teboul2011shape}
Teboul, O.; Kokkinos, I.; Simon, L.; Koutsourakis, P.; and Paragios, N. 2011.
\newblock Shape grammar parsing via reinforcement learning.
\newblock In \emph{Proceedings of the IEEE Conference on Computer Vision and
  Pattern Recognition}, 2273--2280. IEEE.

\bibitem[{Tyle{\v{c}}ek and {\v{S}}{\'a}ra(2013)}]{tylevcek2013spatial}
Tyle{\v{c}}ek, R.; and {\v{S}}{\'a}ra, R. 2013.
\newblock Spatial pattern templates for recognition of objects with regular
  structure.
\newblock In \emph{German Conference on Pattern Recognition}, 364--374.
  Springer.

\bibitem[{Wegner et~al.(2016)Wegner, Branson, Hall, Schindler, and
  Perona}]{DBLP:conf/cvpr/WegnerBHSP16}
Wegner, J.~D.; Branson, S.; Hall, D.; Schindler, K.; and Perona, P. 2016.
\newblock Cataloging public objects using aerial and street-level images-urban
  trees.
\newblock In \emph{Proceedings of the IEEE Conference on Computer Vision and
  Pattern Recognition}, 6014--6023.

\bibitem[{Workman, Souvenir, and Jacobs(2015)}]{workman2015wide}
Workman, S.; Souvenir, R.; and Jacobs, N. 2015.
\newblock Wide-area image geolocalization with aerial reference imagery.
\newblock In \emph{Proceedings of the IEEE International Conference on Computer
  Vision}, 3961--3969.

\bibitem[{Workman et~al.(2017)Workman, Zhai, Crandall, and
  Jacobs}]{DBLP:journals/corr/abs-1708-03035}
Workman, S.; Zhai, M.; Crandall, D.~J.; and Jacobs, N. 2017.
\newblock A Unified Model for Near and Remote Sensing.
\newblock In \emph{Proceedings of the IEEE International Conference on Computer
  Vision}, 2688--2697.

\bibitem[{Wu, Frahm, and Pollefeys(2010)}]{wu2010detecting}
Wu, C.; Frahm, J.-M.; and Pollefeys, M. 2010.
\newblock Detecting large repetitive structures with salient boundaries.
\newblock In \emph{Proceedings of the European Conference on Computer Vision},
  142--155. Springer.

\bibitem[{Yang and F{\"o}rstner(2011)}]{yang2011hierarchical}
Yang, M.~Y.; and F{\"o}rstner, W. 2011.
\newblock A hierarchical conditional random field model for labeling and
  classifying images of man-made scenes.
\newblock In \emph{2011 IEEE International Conference on Computer Vision
  Workshops (ICCV Workshops)}, 196--203. IEEE.

\bibitem[{Yao et~al.(2018)Yao, Luo, Li, Fang, and
  Quan}]{DBLP:conf/eccv/YaoLLFQ18}
Yao, Y.; Luo, Z.; Li, S.; Fang, T.; and Quan, L. 2018.
\newblock Mvsnet: Depth inference for unstructured multi-view stereo.
\newblock In \emph{Proceedings of the European Conference on Computer Vision},
  767--783.

\bibitem[{Zeppelzauer et~al.(2018)Zeppelzauer, Despotovic, Sakeena, Koch, and
  D{\"o}ller}]{zeppelzauer2018automatic}
Zeppelzauer, M.; Despotovic, M.; Sakeena, M.; Koch, D.; and D{\"o}ller, M.
  2018.
\newblock Automatic prediction of building age from photographs.
\newblock In \emph{Proceedings of the 2018 ACM on International Conference on
  Multimedia Retrieval}, 126--134. ACM.

\bibitem[{Zhai et~al.(2017)Zhai, Bessinger, Workman, and
  Jacobs}]{zhai2017predicting}
Zhai, M.; Bessinger, Z.; Workman, S.; and Jacobs, N. 2017.
\newblock Predicting ground-level scene layout from aerial imagery.
\newblock In \emph{Proceedings of the IEEE Conference on Computer Vision and
  Pattern Recognition}, 867--875.

\bibitem[{Zhang et~al.(2013)Zhang, Xu, Jiang, Lin, Cohen-Or, and
  Chen}]{zhang2013layered}
Zhang, H.; Xu, K.; Jiang, W.; Lin, J.; Cohen-Or, D.; and Chen, B. 2013.
\newblock Layered analysis of irregular facades via symmetry maximization.
\newblock \emph{ACM Transactions on Graphics (TOG)} 32(4): 121--1.

\bibitem[{Zhu, Deng, and Newsam(2019)}]{zhu2019fine}
Zhu, Y.; Deng, X.; and Newsam, S. 2019.
\newblock Fine-grained land use classification at the city scale using
  ground-level images.
\newblock \emph{IEEE Transactions on Multimedia} .

\end{thebibliography}


\begin{thebibliography}{16}
\providecommand{\natexlab}[1]{#1}
\providecommand{\url}[1]{\texttt{#1}}
\providecommand{\urlprefix}{URL }
\expandafter\ifx\csname urlstyle\endcsname\relax
  \providecommand{\doi}[1]{doi:\discretionary{}{}{}#1}\else
  \providecommand{\doi}{doi:\discretionary{}{}{}\begingroup
  \urlstyle{rm}\Url}\fi

\bibitem[{Albert, Kaur, and Gonzalez(2017)}]{albert2017using}
Albert, A.; Kaur, J.; and Gonzalez, M.~C. 2017.
\newblock Using convolutional networks and satellite imagery to identify
  patterns in urban environments at a large scale.
\newblock In \emph{Proceedings of the 23rd ACM SIGKDD international conference
  on knowledge discovery and data mining}, 1357--1366.

\bibitem[{Basu et~al.(2015)Basu, Ganguly, Mukhopadhyay, DiBiano, Karki, and
  Nemani}]{basu2015deepsat}
Basu, S.; Ganguly, S.; Mukhopadhyay, S.; DiBiano, R.; Karki, M.; and Nemani, R.
  2015.
\newblock Deepsat: a learning framework for satellite imagery.
\newblock In \emph{Proceedings of the 23rd SIGSPATIAL international conference
  on advances in geographic information systems}, 1--10.

\bibitem[{B{\'o}dis-Szomor{\'u}, Riemenschneider, and
  Van~Gool(2016)}]{bodis2016efficient}
B{\'o}dis-Szomor{\'u}, A.; Riemenschneider, H.; and Van~Gool, L. 2016.
\newblock Efficient volumetric fusion of airborne and street-side data for
  urban reconstruction.
\newblock In \emph{International Conference on Pattern Recognition},
  3204--3209. IEEE.

\bibitem[{Frohlich, Rodner, and Denzler(2010)}]{frohlich2010fast}
Frohlich, B.; Rodner, E.; and Denzler, J. 2010.
\newblock A fast approach for pixelwise labeling of facade images.
\newblock In \emph{2010 20th International Conference on Pattern Recognition},
  3029--3032. IEEE.

\bibitem[{Gadde, Marlet, and Paragios(2016)}]{gadde2016learning}
Gadde, R.; Marlet, R.; and Paragios, N. 2016.
\newblock Learning grammars for architecture-specific facade parsing.
\newblock \emph{International Journal of Computer Vision} 117(3): 290--316.

\bibitem[{Kang et~al.(2018)Kang, K{\"o}rner, Wang, Taubenb{\"o}ck, and
  Zhu}]{kang2018building}
Kang, J.; K{\"o}rner, M.; Wang, Y.; Taubenb{\"o}ck, H.; and Zhu, X.~X. 2018.
\newblock Building instance classification using street view images.
\newblock \emph{ISPRS Journal of Photogrammetry and Remote Sensing} .

\bibitem[{Kor{\v c} and F{\" o}rstner(2009)}]{korc}
Kor{\v c}, F.; and F{\" o}rstner, W. 2009.
\newblock {eTRIMS} {I}mage {D}atabase for Interpreting Images of Man-Made
  Scenes.
\newblock Technical Report TR-IGG-P-2009-01, Dept. of Photogrammetry,
  University of Bonn.

\bibitem[{Lotte et~al.(2018)Lotte, Haala, Karpina, Arag{\~a}o, Shimabukuro
  et~al.}]{lotte20183d}
Lotte, R.; Haala, N.; Karpina, M.; Arag{\~a}o, L.; Shimabukuro, Y.; et~al.
  2018.
\newblock 3D Fa{\c{c}}ade Labeling over Complex Scenarios: A Case Study Using
  Convolutional Neural Network and Structure-From-Motion.
\newblock \emph{Remote Sensing} .

\bibitem[{Riemenschneider et~al.(2014)Riemenschneider, B{\'o}dis-Szomor{\'u},
  Weissenberg, and Van~Gool}]{riemenschneider2014learning}
Riemenschneider, H.; B{\'o}dis-Szomor{\'u}, A.; Weissenberg, J.; and Van~Gool,
  L. 2014.
\newblock Learning where to classify in multi-view semantic segmentation.
\newblock In \emph{Proceedings of the European Conference on Computer Vision},
  516--532. Springer.

\bibitem[{Riemenschneider et~al.(2012)Riemenschneider, Krispel, Thaller,
  Donoser, Havemann, Fellner, and Bischof}]{riemenschneider2012irregular}
Riemenschneider, H.; Krispel, U.; Thaller, W.; Donoser, M.; Havemann, S.;
  Fellner, D.; and Bischof, H. 2012.
\newblock Irregular lattices for complex shape grammar facade parsing.
\newblock In \emph{Proceedings of the IEEE Conference on Computer Vision and
  Pattern Recognition}, 1640--1647. IEEE.

\bibitem[{Shao, Svoboda, and Van~Gool(2003)}]{shao2003zubud}
Shao, H.; Svoboda, T.; and Van~Gool, L. 2003.
\newblock Zubud-zurich buildings database for image based recognition.
\newblock \emph{Computer Vision Lab, Swiss Federal Institute of Technology,
  Switzerland, Tech. Rep} 260(20): 6.

\bibitem[{Teboul et~al.(2011)Teboul, Kokkinos, Simon, Koutsourakis, and
  Paragios}]{teboul2011shape}
Teboul, O.; Kokkinos, I.; Simon, L.; Koutsourakis, P.; and Paragios, N. 2011.
\newblock Shape grammar parsing via reinforcement learning.
\newblock In \emph{Proceedings of the IEEE Conference on Computer Vision and
  Pattern Recognition}, 2273--2280. IEEE.

\bibitem[{Tyle{\v{c}}ek and {\v{S}}{\'a}ra(2013)}]{tylevcek2013spatial}
Tyle{\v{c}}ek, R.; and {\v{S}}{\'a}ra, R. 2013.
\newblock Spatial pattern templates for recognition of objects with regular
  structure.
\newblock In \emph{German Conference on Pattern Recognition}, 364--374.
  Springer.

\bibitem[{Workman, Souvenir, and Jacobs(2015)}]{workman2015wide}
Workman, S.; Souvenir, R.; and Jacobs, N. 2015.
\newblock Wide-area image geolocalization with aerial reference imagery.
\newblock In \emph{Proceedings of the IEEE International Conference on Computer
  Vision}, 3961--3969.

\bibitem[{Workman et~al.(2017)Workman, Zhai, Crandall, and
  Jacobs}]{DBLP:journals/corr/abs-1708-03035}
Workman, S.; Zhai, M.; Crandall, D.~J.; and Jacobs, N. 2017.
\newblock A Unified Model for Near and Remote Sensing.
\newblock In \emph{Proceedings of the IEEE International Conference on Computer
  Vision}, 2688--2697.

\bibitem[{Yang and Newsam(2010)}]{yang2010bag}
Yang, Y.; and Newsam, S. 2010.
\newblock Bag-of-visual-words and spatial extensions for land-use
  classification.
\newblock In \emph{Proceedings of the 18th SIGSPATIAL international conference
  on advances in geographic information systems}, 270--279.

\end{thebibliography}

\end{document}


\onecolumn
\appendix

\section{Building location heatmap}

\begin{figure}[h]
\centering
\includegraphics[width=\linewidth]{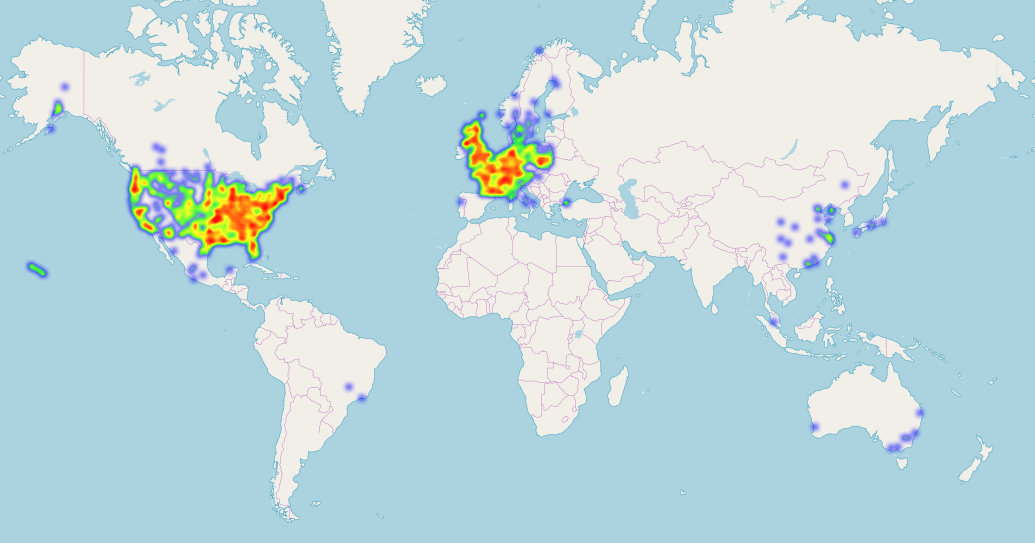}
\caption[A world heatmap of the building locations.]{A world heatmap of the building locations for our dataset. Our dataset represents a broad variety of buildings and architectural styles.}
\end{figure}

\section{Resolution of the top-view imagery}\label{appendix:resolution}
Aerial photos' spatial resolution depends on the latitude and can be expressed as \textit{zoom $19$} using the \textit{slippy map} definition. The exact spatial resolution $r$, measured in meters per pixel, is expressed by the following formula:
$$r = \text{equator length} \cdot \cos(\text{latitude}) \cdot \frac{1}{2^{19}}\cdot \frac{1}{2^{8}}.$$
The factor of $2^{-19}$ corresponds to the zoom, whereas the factor of $2^{-8}$ matches with the square tile size of $256$ pixels.

\section{Dataset details}

In this section, we provide more dataset examples, as well as a comparison with existing datasets from the literature.

In Table~\ref{tab:examples}, we show one instance of every class for every attribute. In Figures~2-7, we show challenging scenes from our dataset, together with descriptions explaining how correct classification is still possible. Finally, in Table~\ref{tab:comparison} we compare our dataset with existing datasets in the same domain.

\begin{table*}[h!]
\centering
\resizebox{\textwidth}{!}{%
\begin{tabular}{| c | c c c c c c |}
\hline
\addvbuffer[0ex -1.5ex]{\rot{Construction type}} &
\addvbuffer[1ex 0ex]{\includegraphics[width=0.14\linewidth]{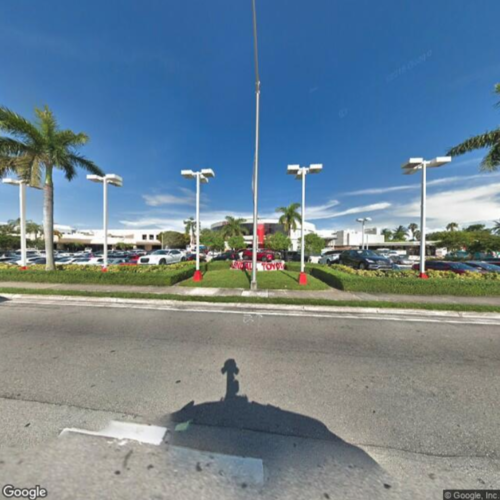}} &
\includegraphics[width=0.14\linewidth]{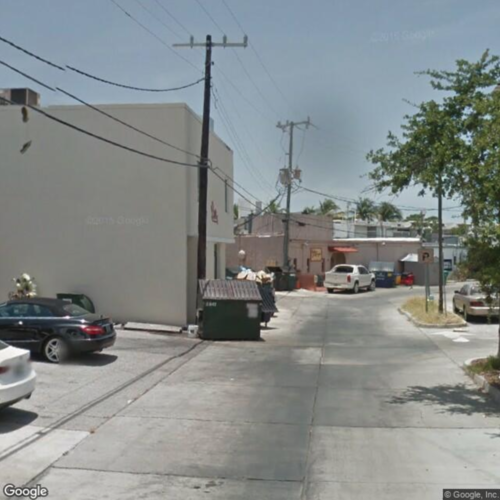} &
\includegraphics[width=0.14\linewidth]{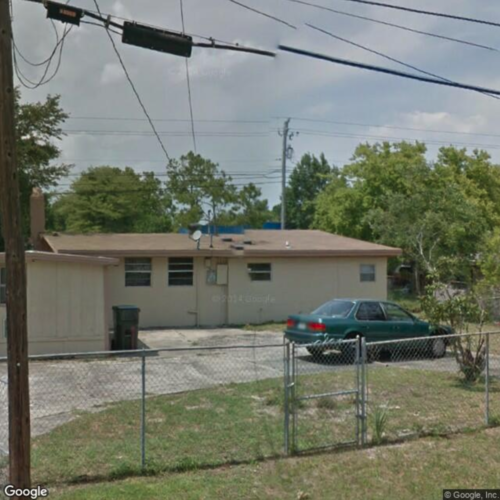} &
\includegraphics[width=0.14\linewidth]{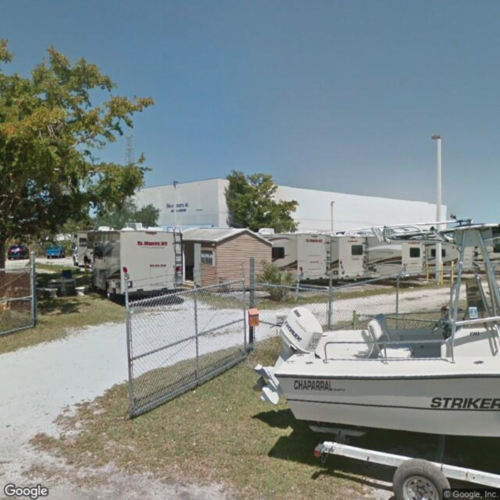} & &\\
& Concrete (5071) & Masonry (2230) & Wood (1195) & Metal (1036) & & \\
\hline
\addvbuffer[0ex 2.5ex]{\rot{Roof pitch}} &
\addvbuffer[1ex 0ex]{\includegraphics[width=0.14\linewidth]{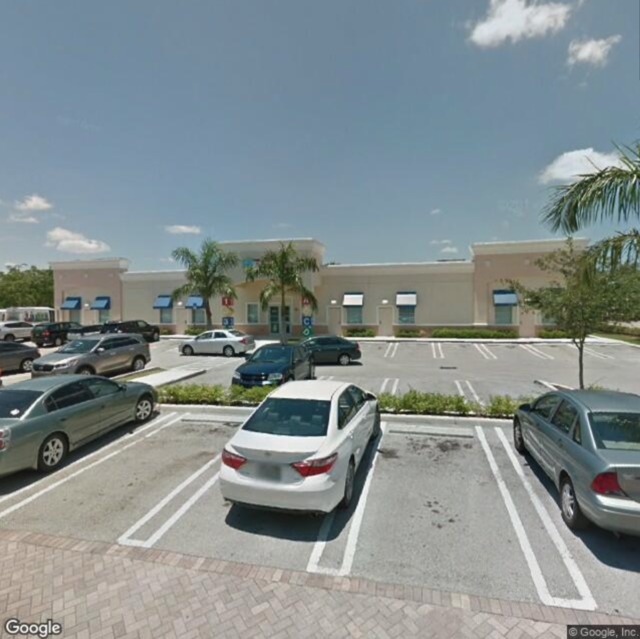}} &
\includegraphics[width=0.14\linewidth]{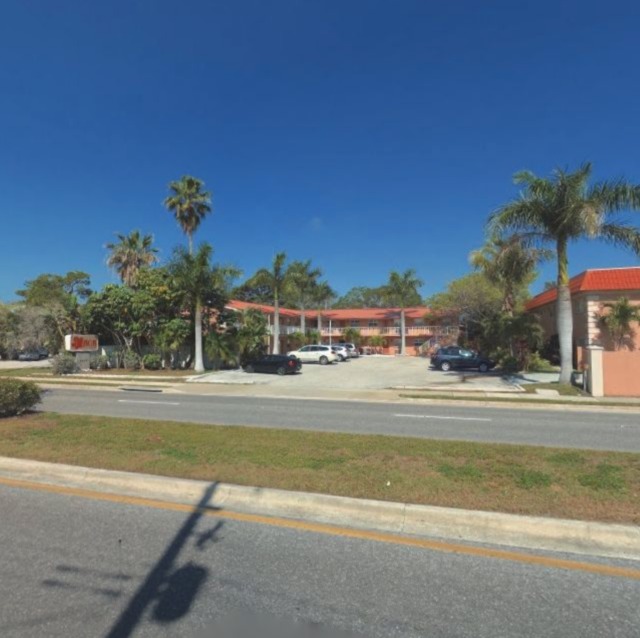} &
\includegraphics[width=0.14\linewidth]{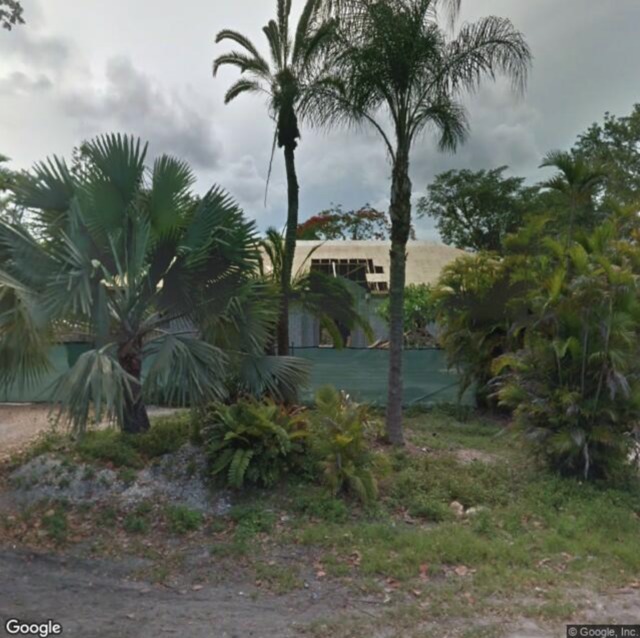} & 
\includegraphics[width=0.14\linewidth]{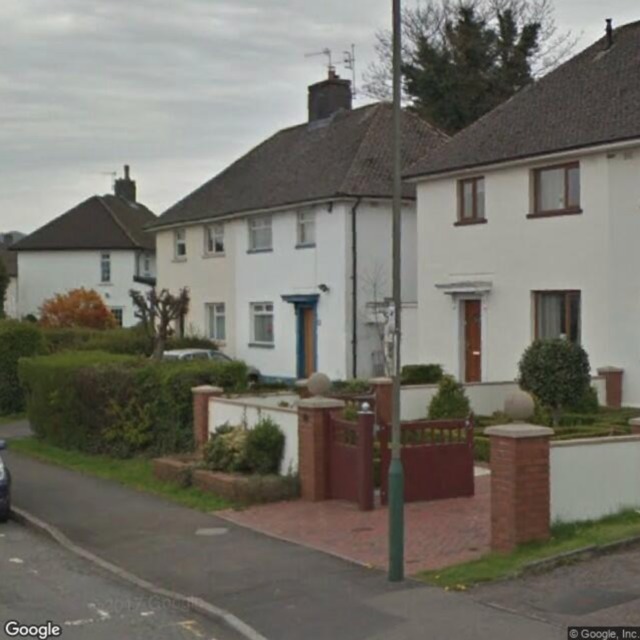} & & \\  
& Flat (5382) & Low (1849) & Moderate (2232) & High (214) & & \\   
\hline
\addvbuffer[0ex 0ex]{\rot{Roof geometry}} &
\addvbuffer[1ex 0ex]{\includegraphics[width=0.14\linewidth]{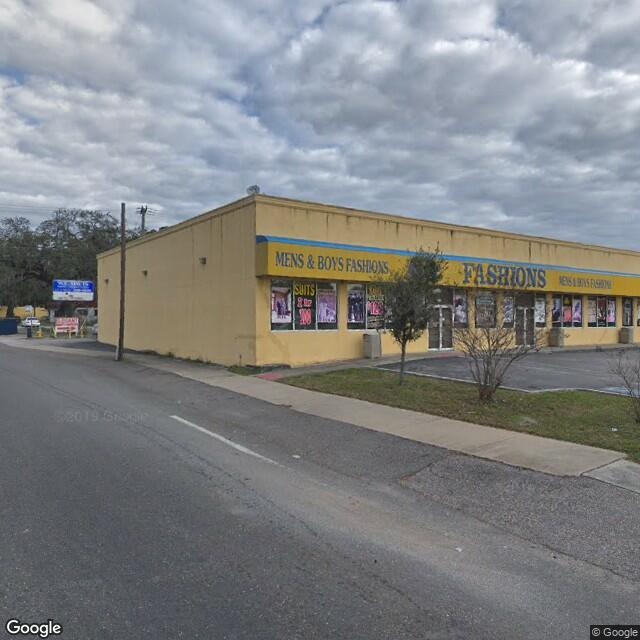}} & 
\includegraphics[width=0.14\linewidth]{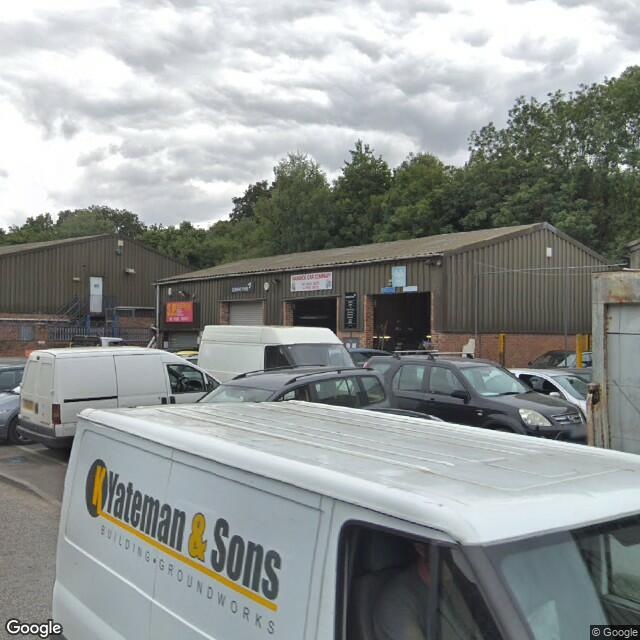} & 
\includegraphics[width=0.14\linewidth]{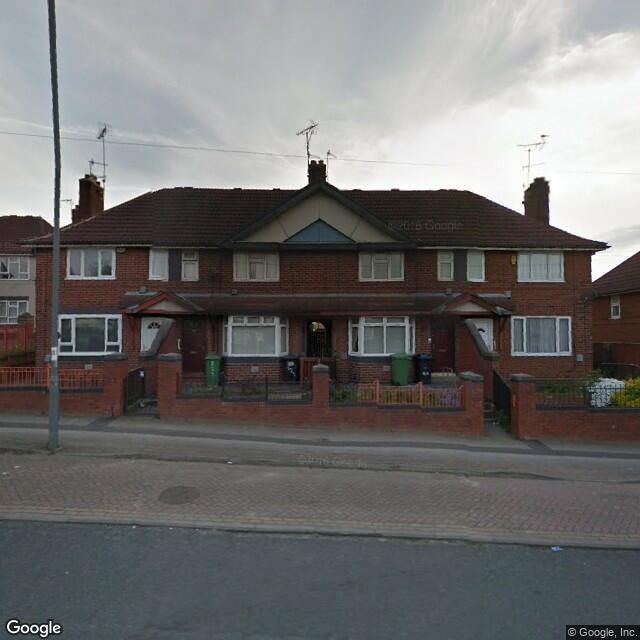} & 
\includegraphics[width=0.14\linewidth]{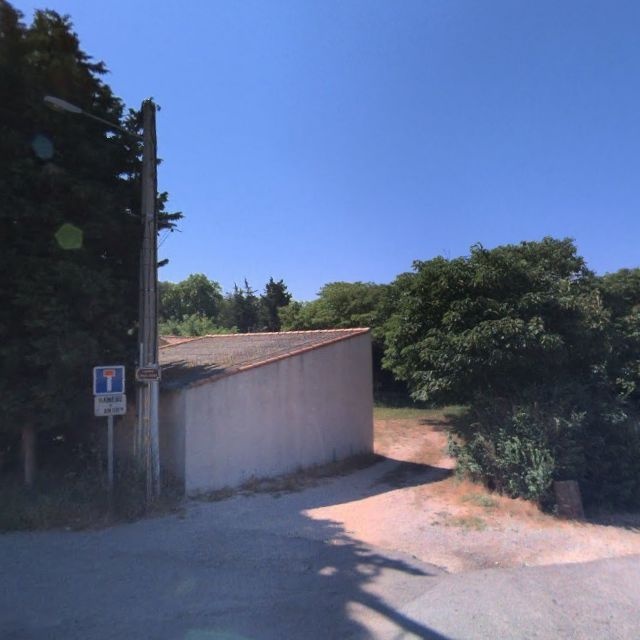} & & \\
& Flat (4933) & Gable (1532) & Hip (553) & Shed (74) & & \\
\hline
\addvbuffer[0ex -0.5ex]{\rot{Occupancy type}} &
\addvbuffer[1ex 0ex]{\includegraphics[width=0.14\linewidth]{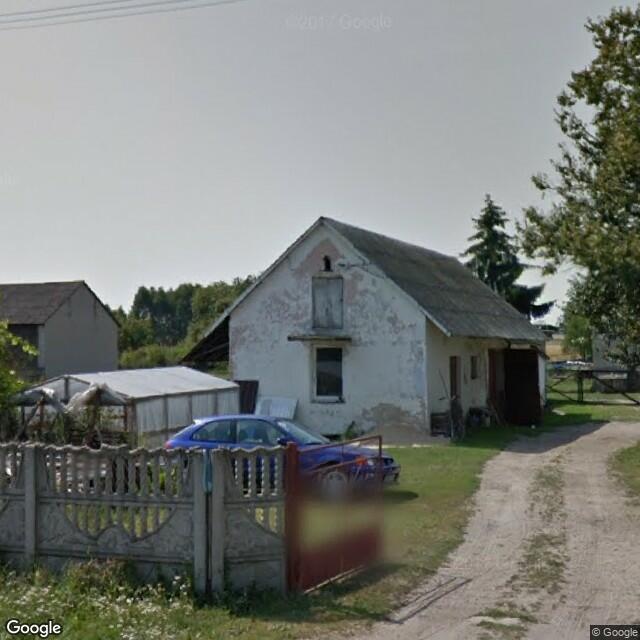}} & 
\includegraphics[width=0.14\linewidth]{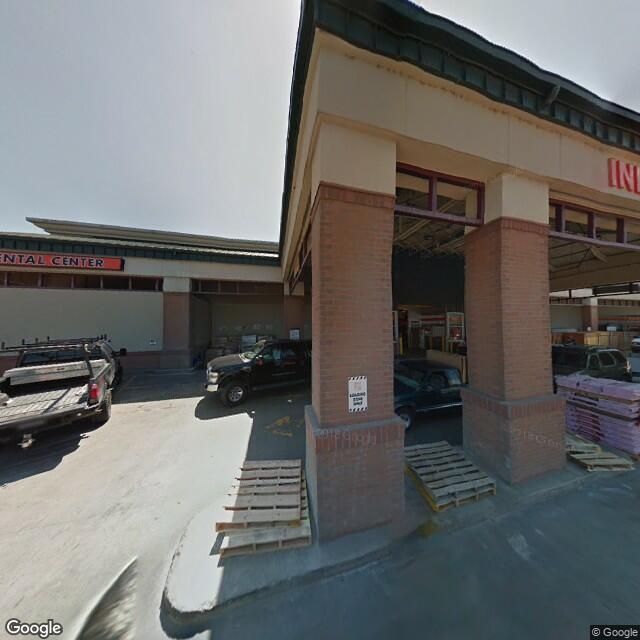} &
\includegraphics[width=0.14\linewidth]{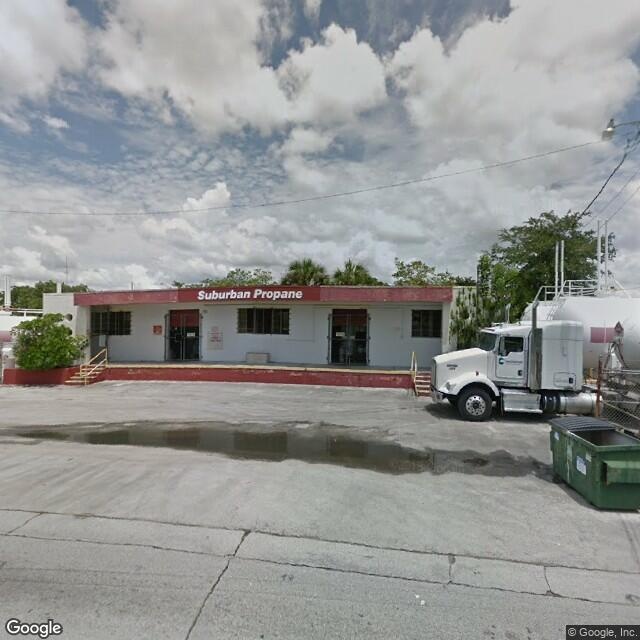} &
\includegraphics[width=0.14\linewidth]{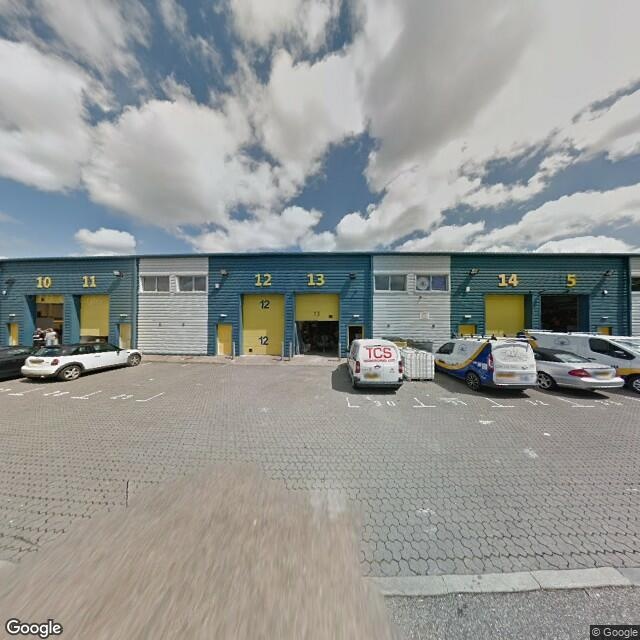} &
\includegraphics[width=0.14\linewidth]{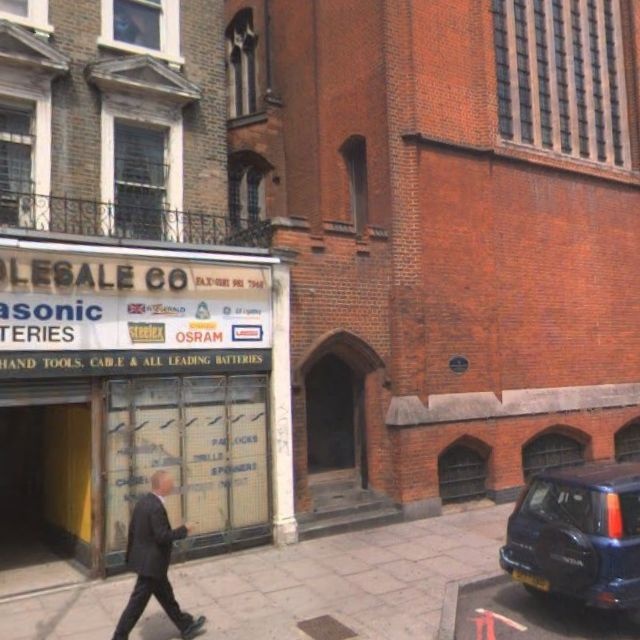} &
\includegraphics[width=0.14\linewidth]{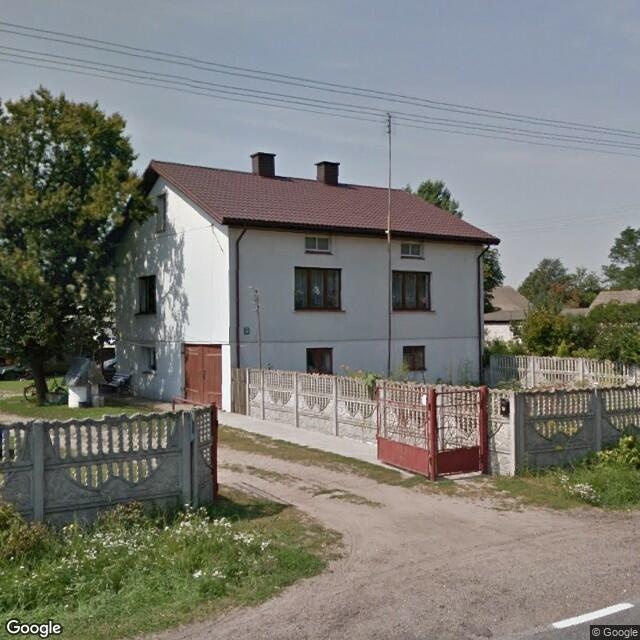} \\
& Agriculture (89) & Commercial (4224) & Industrial (714) & Mercantile (253) & Public (389) & Residential (2475)\\
\hline
\addvbuffer[0ex -1ex]{\rot{Number of floors}} & 
\addvbuffer[1ex 0ex]{\includegraphics[width=0.14\linewidth]{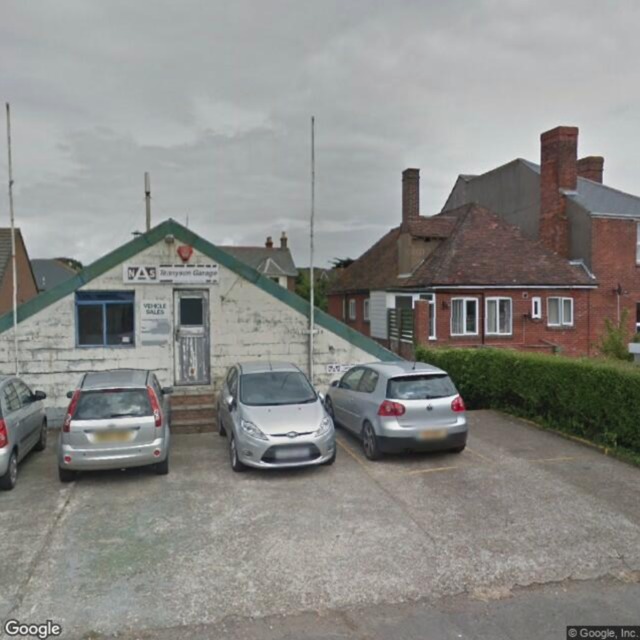}} & 
\includegraphics[width=0.14\linewidth]{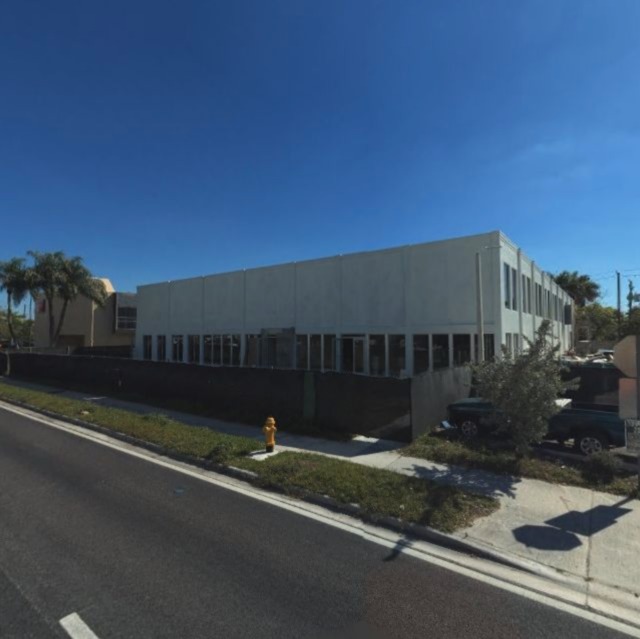} & 
\includegraphics[width=0.14\linewidth]{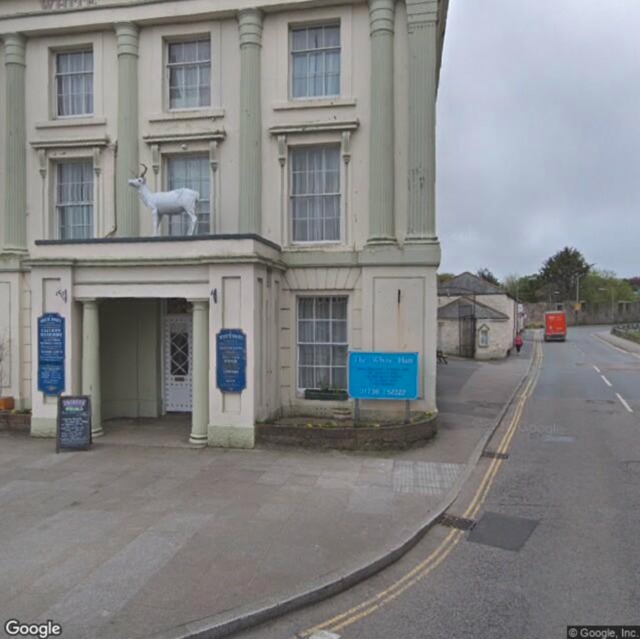} & 
\includegraphics[width=0.14\linewidth]{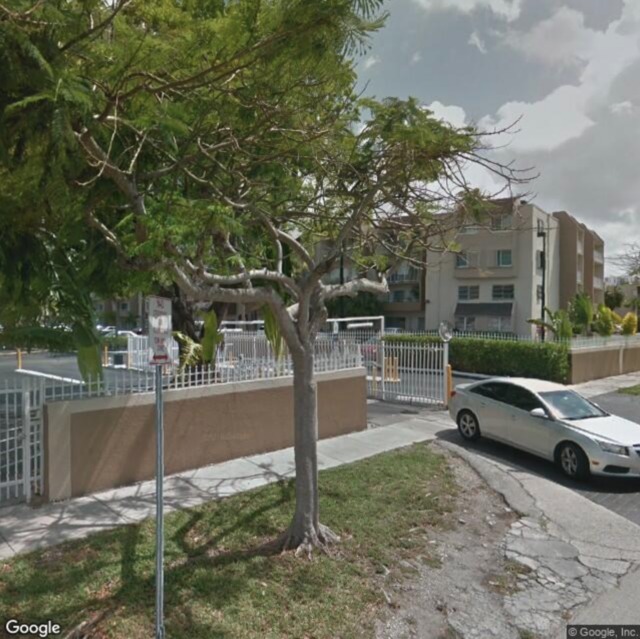} &
\includegraphics[width=0.14\linewidth]{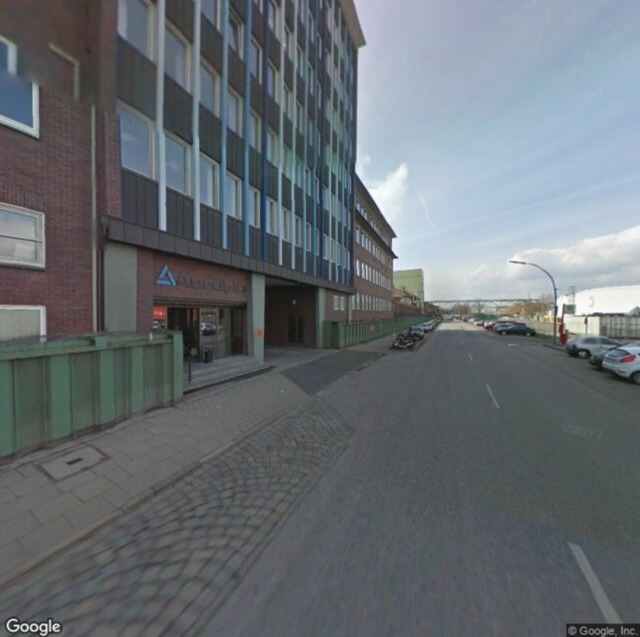} & \\
& 1 (4438) & 2 (2864) & 3 (873) & 4 (390) & 5 or more (1067) & \\
\hline
\addvbuffer[0ex 2.5ex]{\multirow{2}{*}{\rot{Facade material}}} & 
\addvbuffer[1ex 0ex]{\includegraphics[width=0.14\linewidth]{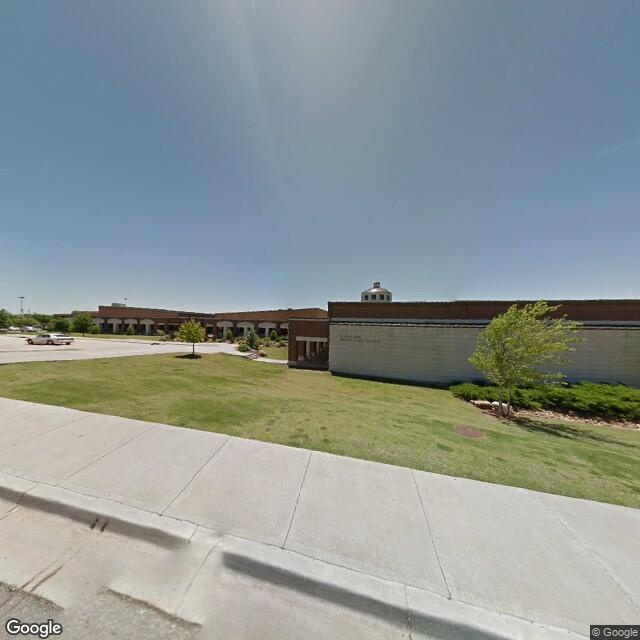}} & 
\includegraphics[width=0.14\linewidth]{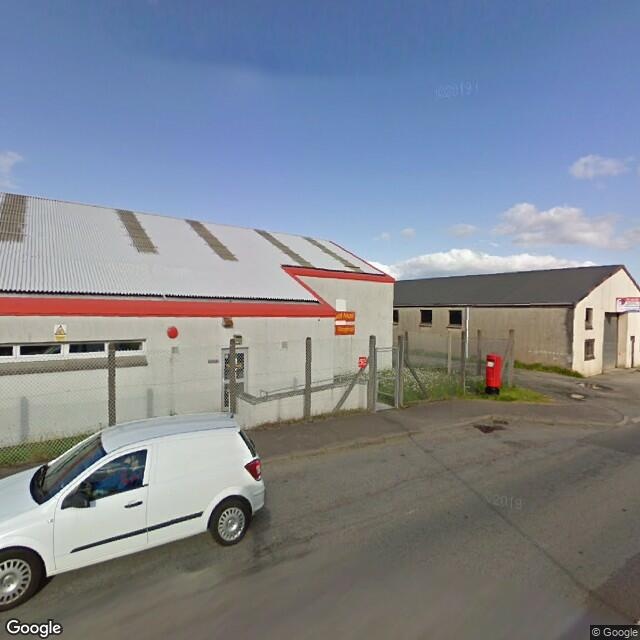} & 
\includegraphics[width=0.14\linewidth]{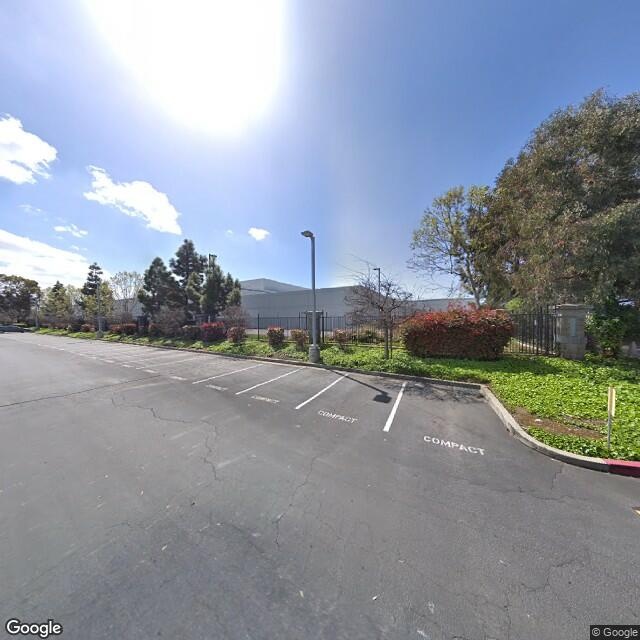} & 
\includegraphics[width=0.14\linewidth]{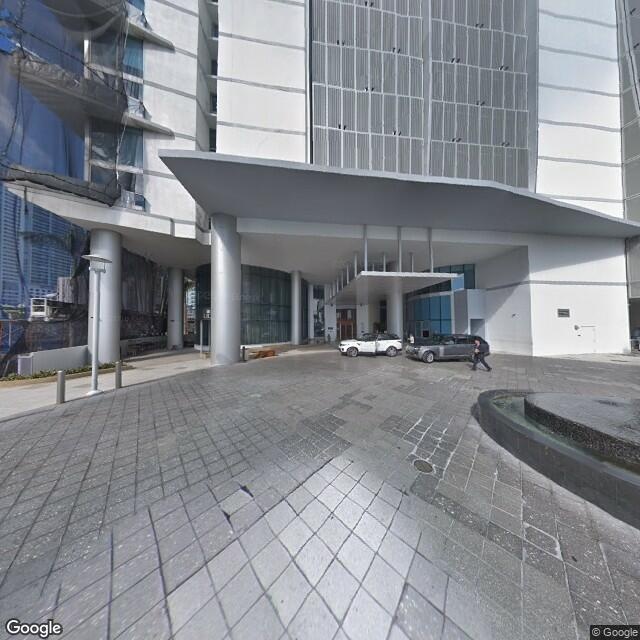} &
\includegraphics[width=0.14\linewidth]{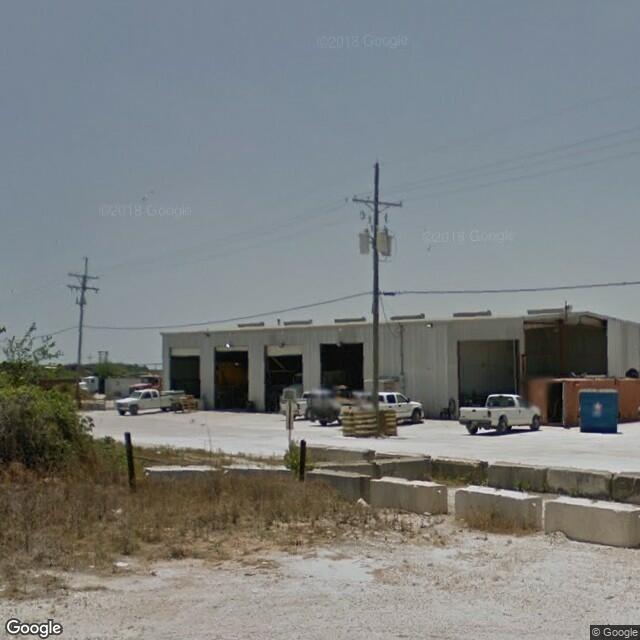} & \\
& Brick (1274) & Cement block (403) & Concrete (1182) & Glass (86) & Metal (829) & \\
& 
\includegraphics[width=0.14\linewidth]{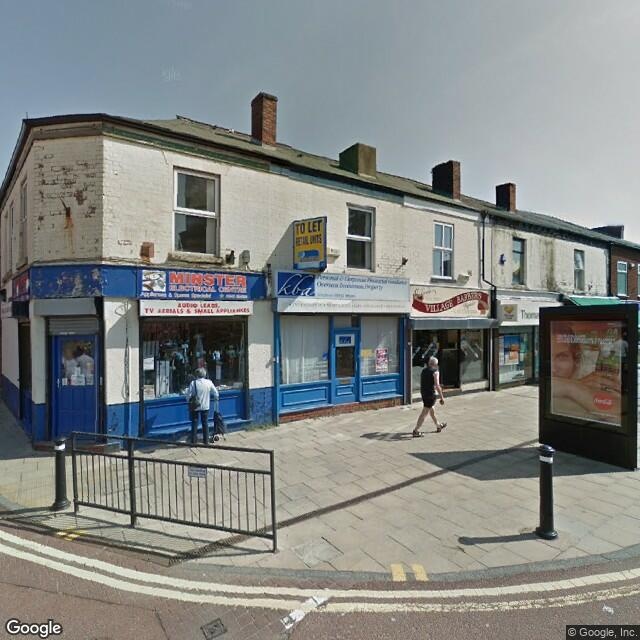} & 
\includegraphics[width=0.14\linewidth]{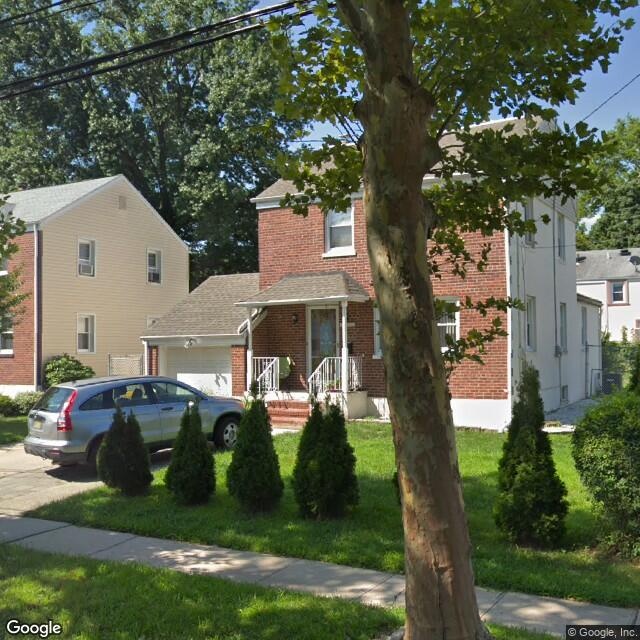} & 
\includegraphics[width=0.14\linewidth]{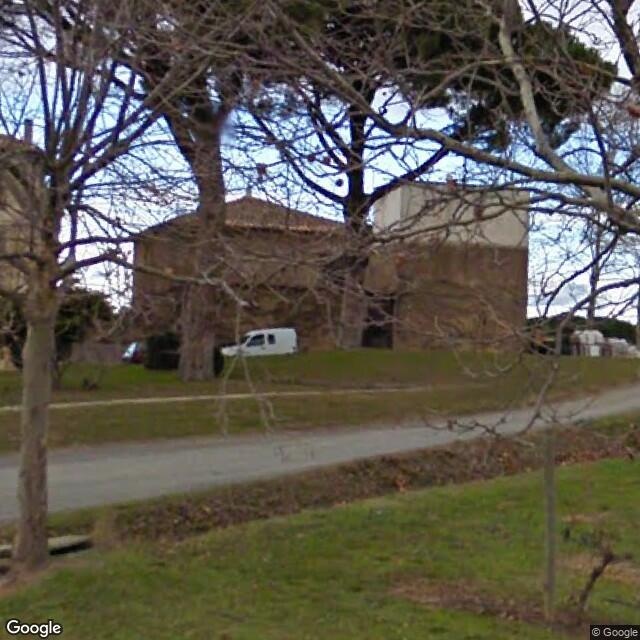} & 
\includegraphics[width=0.14\linewidth]{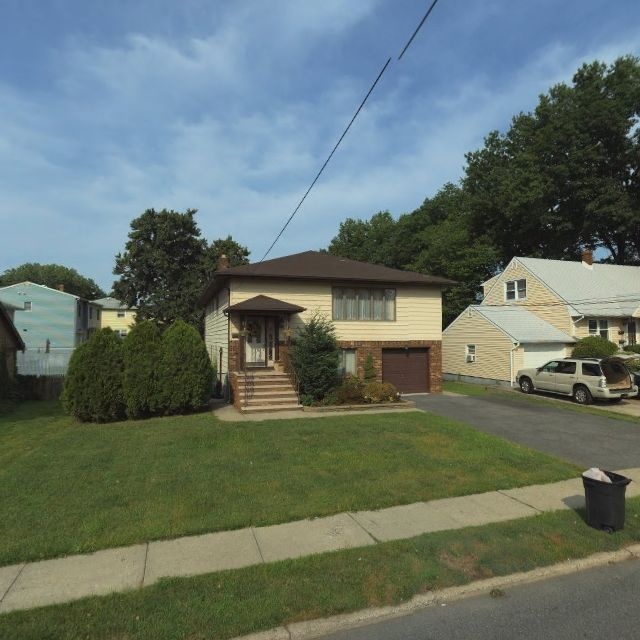} & & \\
& Plaster (3253) & Plastic (359) & Stone (119) & Wood (133) & & \\
\hline
\end{tabular}
}
\caption[Examples of street-view images for every class.]{Examples of street-view images for every class in the dataset together with the building counter for that class in the training dataset. The dataset presents many challenges, such as occlusions and small object recognition from afar. Some images contain multiple buildings, confusing the model, and a few of them even miss the correct building.}
\label{tab:examples}
\end{table*}

\clearpage

\begin{figure*}[h!]
\centering
\includegraphics[width=0.9\linewidth]{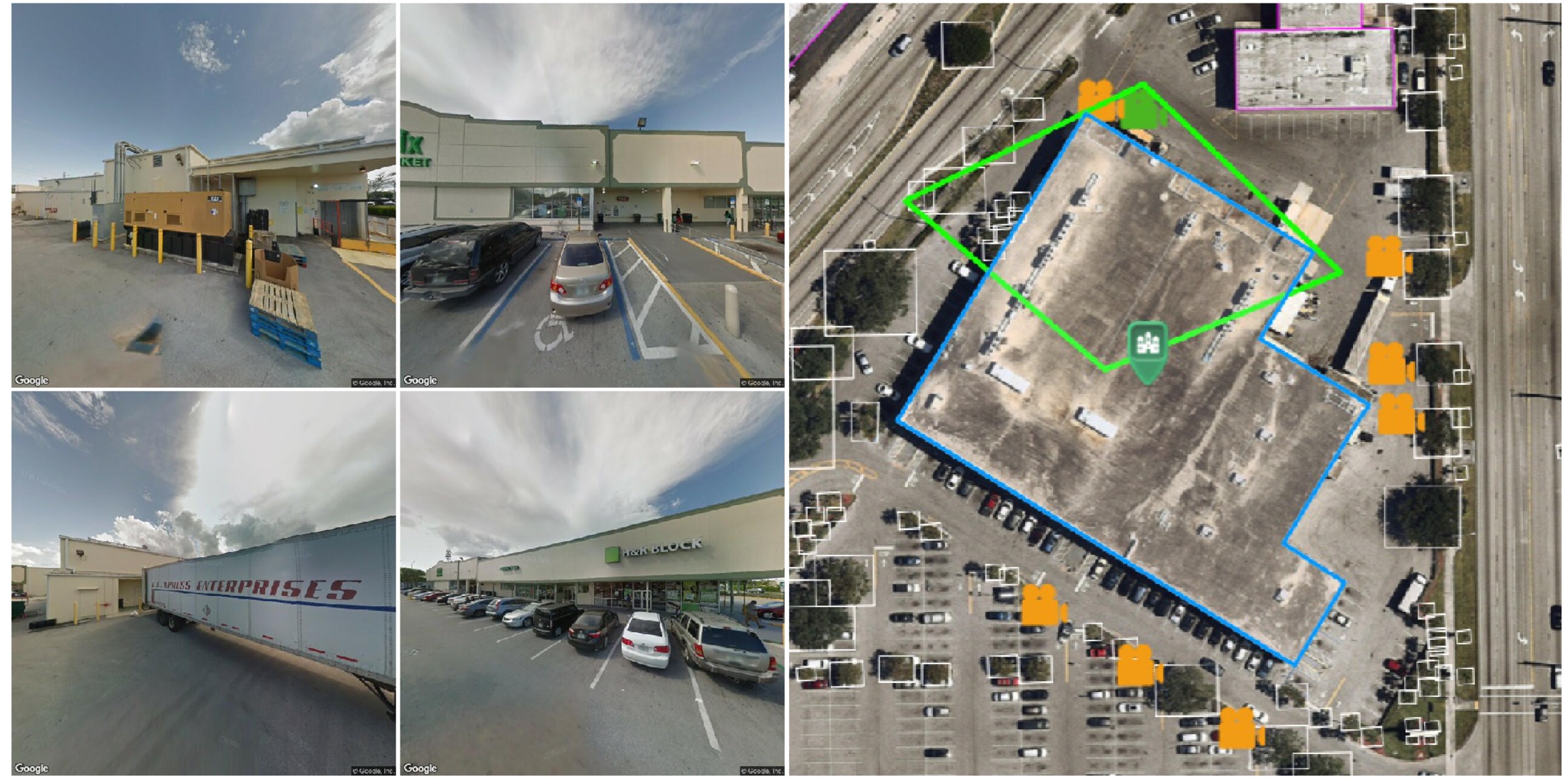}
\caption[Difficult example 1.]{Photos of a shopping mall from completely different perspectives. The model sees three different sides of the building from different angles and must be able to filter out the cars appearing in images.}
\label{fig:ex_1}
\end{figure*}

\begin{figure*}[h!]
\centering
\includegraphics[width=0.9\linewidth]{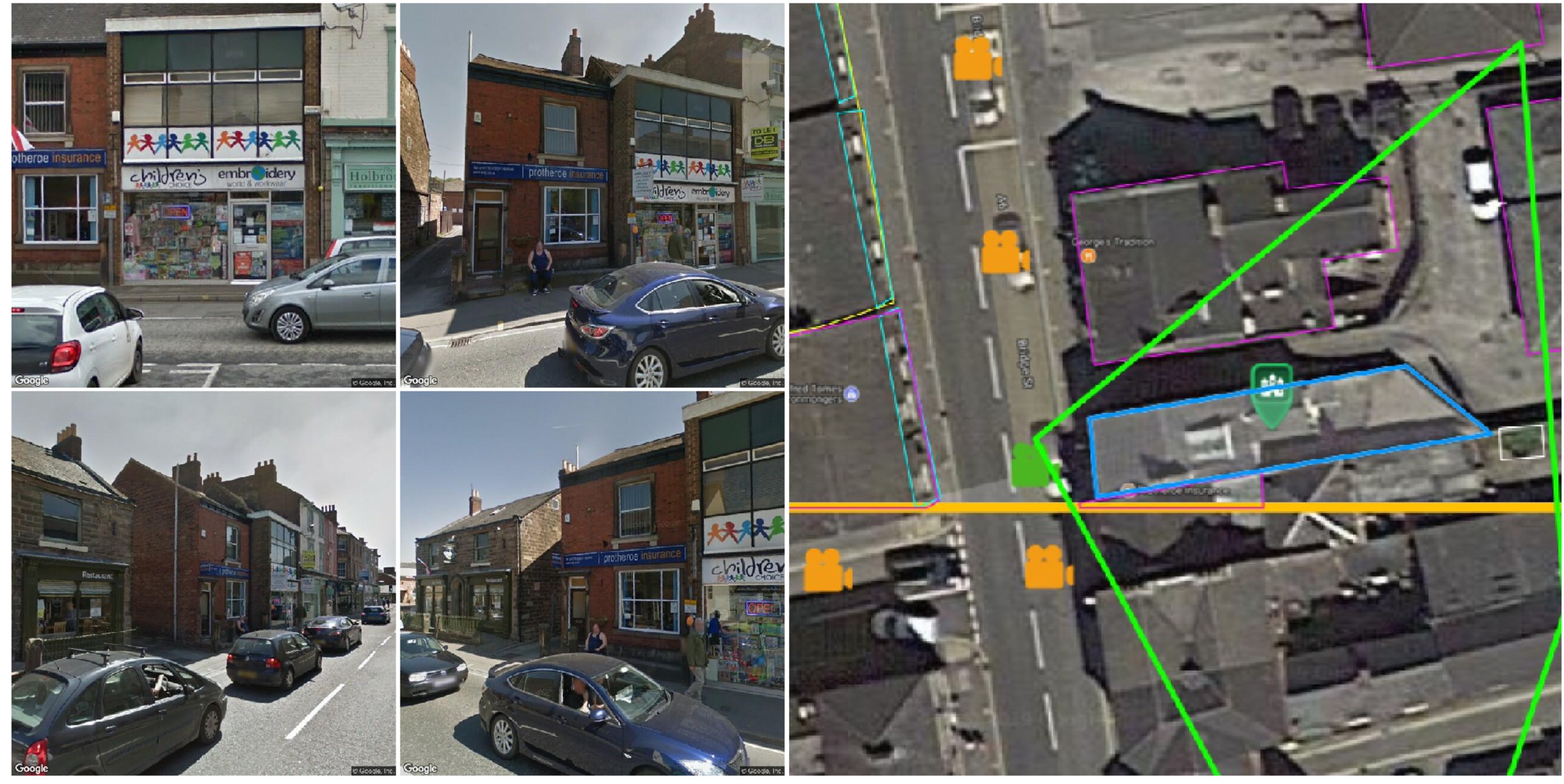}
\caption[Difficult example 2.]{A typical example from an urban area, where more than one building is visible on the photo. Moreover, it is not ideally cropped in the second image. Observe that the roof pitch could be correctly estimated only from the third photo.}
\label{fig:ex_3}
\end{figure*}

\clearpage

\begin{figure*}[t]
\centering
\includegraphics[width=0.9\linewidth]{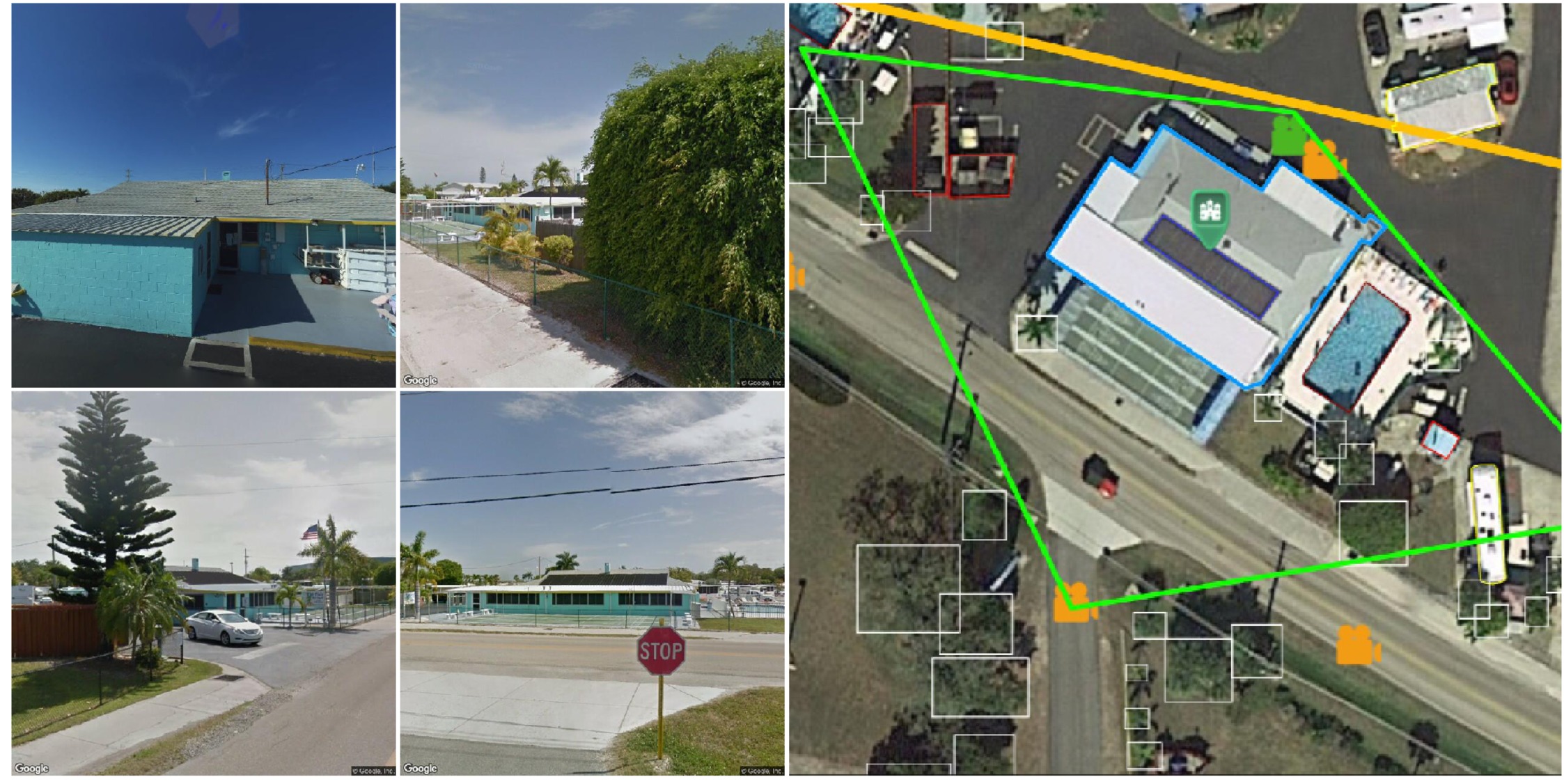}
\caption[Difficult example 3.]{Different photos of the same building highlighting various difficulties of the dataset. The first image is from a very close distance. The second image is partially occluded; the third one is also occluded by a visible car. The last one is from afar. Note that we can infer the construction type only from the first photo, where the bricks can be seen, whereas the last photo clearly shows the number of stories.}
\label{fig:ex_2}
\end{figure*}

\begin{figure*}[b]
\centering
\includegraphics[width=0.9\linewidth]{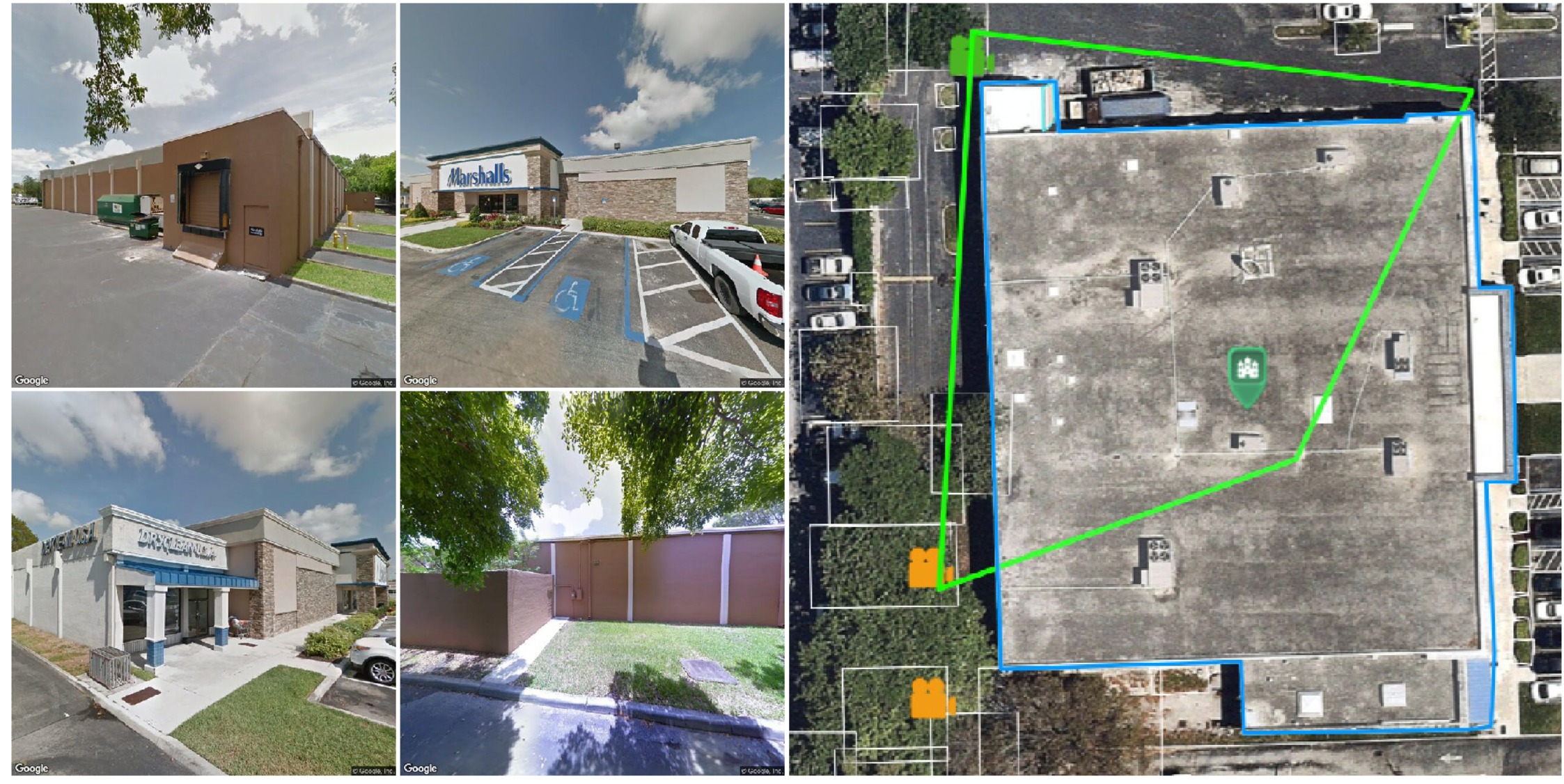}
\caption[Difficult example 4.]{Four different photos of the same building. They all look like different buildings. The last image is partially occluded. The facade on the second and third photo might suggest the building is made of masonry, when the correct answer is reinforced concrete.}
\label{fig:ex_4}
\end{figure*}

\clearpage

\begin{figure*}[t]
\centering
\includegraphics[width=0.9\linewidth]{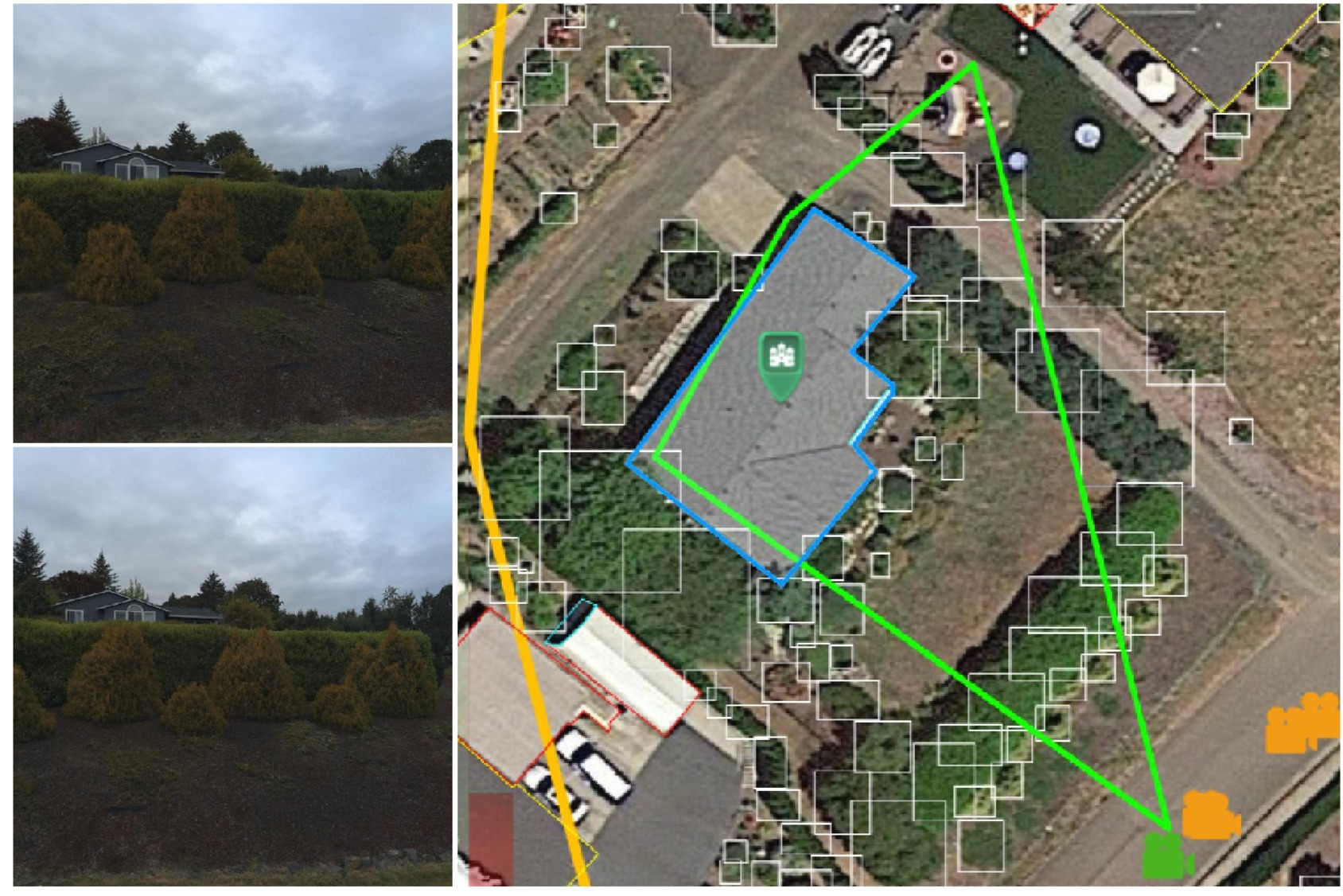}
\caption[Difficult example 5.]{A building which is almost completely occluded but we are still able to infer some of its features. It has one floor, low roof pitch and is made of wood.}
\label{fig:ex_5}
\end{figure*}

\begin{figure*}[b]
\centering
\includegraphics[width=0.9\linewidth]{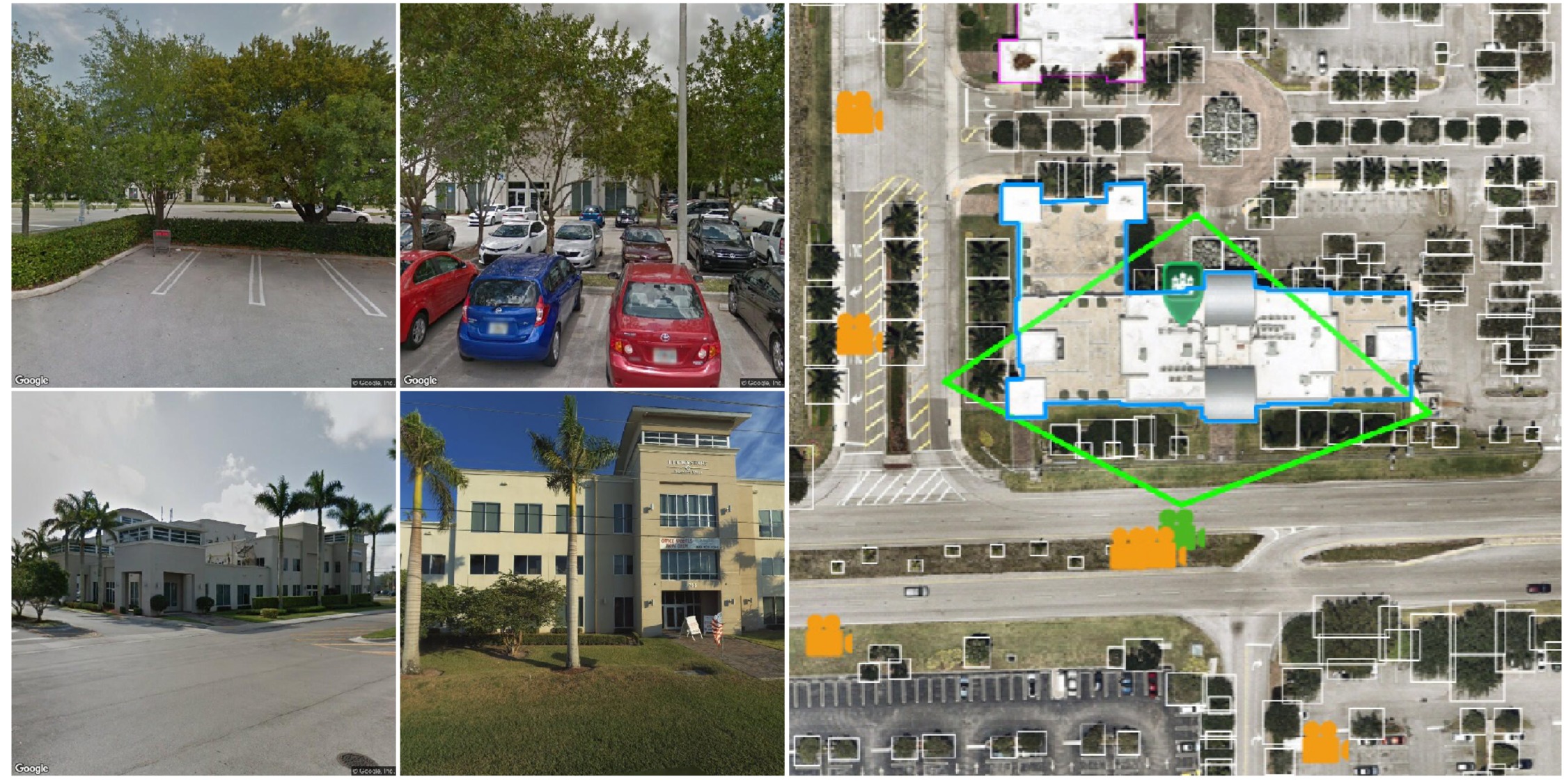}
\caption[Difficult example 6.]{A challenging example, where the building is occluded on the first photo, and partially occluded on the second one with cars present. Moreover, we can infer the number of floors only from the last photo, since the building is not fully visible in the third one. The photos are from different sides of the building, making the task even more difficult.}
\label{fig:ex_6}
\end{figure*}

\clearpage

\begin{table*}
\begin{center}
\caption[Comparison of datasets related to building analysis.]{Comparison of datasets related to building analysis. \#SV - number of street-view images, \#TV - number of top-view images.}
\label{tab:datasets}
\begin{tabular}{lp{1.3cm}p{1.1cm}p{3cm}p{3.5cm}c}
\toprule
Dataset                                             & \#SV & \#TV & Location & Dataset Type & Multi-view \\

\midrule

ICG Graz50 [\citenum{riemenschneider2012irregular}] & 50     & & Graz (Austria) & 2D Segmentation & \ding{55} \\ \midrule
eTRIMS [\citenum{korc}]                                  & 60     & & Multiple & 2D Segmentation & \ding{55} \\ \midrule
ENPC ArtDeco [\citenum{gadde2016learning}]               & 79     & & Paris (France) & 2D Segmentation & \ding{55} \\ \midrule
ECP Hausmannian [\citenum{teboul2011shape}]              & 104    & & Paris (France) & 2D Segmentation & \ding{55} \\ \midrule
CMP [\citenum{tylevcek2013spatial}]                      & 378    & & Multiple & 2D Segmentation & \ding{55} \\ \midrule
LabelMeFacade [\citenum{frohlich2010fast}]               & 945    & & Multiple & 2D Segmentation & \ding{55} \\ \midrule
ZuBuD [\citenum{shao2003zubud}]                          & 1\,005 & & Zurich (Switzerland) & 2D Segmentation & \ding{55} \\ \midrule
RueMonge2014 [\citenum{riemenschneider2014learning}]     & 428    & & France & 2D \& 3D Segmentation & \checkmark \\ \midrule
SJC [\citenum{lotte20183d}]                              & 175    & & Brazil & 2D \& 3D Segmentation & \checkmark \\ \midrule
Limmatquai and Munsterhof [\citenum{bodis2016efficient}] & 1\,476 & 23 & Zurich (Switzerland) & 3D Reconstruction & \checkmark \\ \midrule

CVUSA [\citenum{workman2015wide}]                        & 1\,588\,655 & 879\,318 & USA & Geolocalization & \checkmark \\ \midrule

Kang et al. [\citenum{kang2018building}]                 & 19\,658 & & USA, Canada & Classification \newline (occupancy type) & \ding{55}  \\ \midrule

UC Merced [\citenum{yang2010bag}] & & 2\,100 & USA & Classification (land use) & \ding{55} \\ \midrule
DeepSat [\citenum{basu2015deepsat}] & & 950\,000 & Global & Classification (land use) & \ding{55} \\ \midrule
Albert et al [\citenum{albert2017using}] & & 140\,000 & Europe & Classification (land use) & \ding{55} \\ \midrule

Brooklyn and Queens [\citenum{DBLP:journals/corr/abs-1708-03035}] & 38\,603 & 10\,044 & NYC (USA) & Classification \newline (3 attributes) & \checkmark \\ \midrule 

\textbf{Ours} & \textbf{39\,752} & \textbf{9\,674} & \textbf{Global} & \textbf{Classification \newline (6 attributes)} & \checkmark \\ \bottomrule
\end{tabular}
\label{tab:comparison}
\end{center}
\end{table*}

\section{Baseline fusion}

\begin{figure}[h]
\centering
\includegraphics[width=0.5\linewidth]{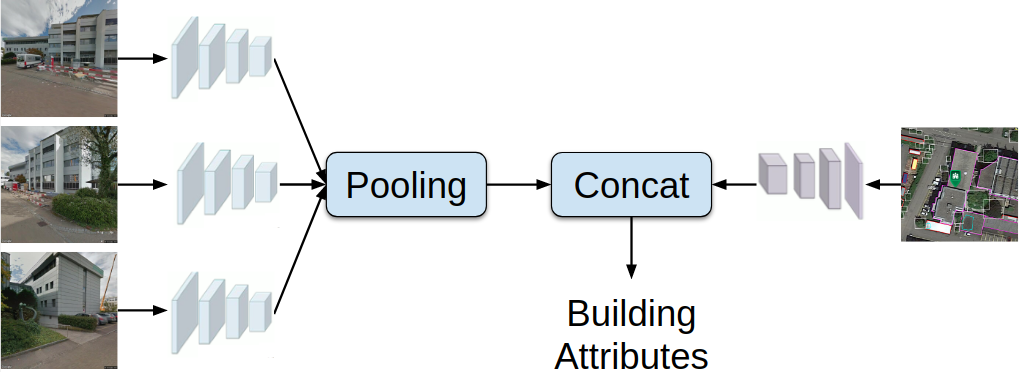}
\caption[Model of the baseline fusion.]{We average the street-view features and then concatenate them with top-view features for the final class prediction.}
\label{fig:baseline_fusion}
\end{figure}

\clearpage

\section{Computing the visible parts of the building}
\label{sec:sweep}
For each street-view image we compute which part of the top-view building polygon is visible. Note that many real-life building outlines are non-convex, and so the visible part might be a disconnected set of pieces of sides of the building polygon. We also take into account the occlusions caused by adjacent buildings. Here, we describe the approaches we investigated, comparing their time complexities in terms of the number of sides of the polygon $n$.

First, let us describe a naive approach that works in $O(n^2)$. We begin by fixing a single side, and computing what part of it is visible (possibly the entire side is occluded, in which case it will be ignored). To check visibility, we simply iterate through all the remaining sides, checking for occlusions.

In practice, we noticed that using a naive $O(n^2)$ algorithm was considerably slowing down both training and inference. This is because the number of polygon vertices -- and thus sides -- in our dataset can be quite large. In particular, one can often find multiple vertices near the building corners, due to post-processing of human annotations for loop closing. Therefore, we also developed a faster sweep line-based algorithm, which works in $O(n \log n)$ time.

\begin{figure}[h]
  \centering
  \includegraphics[width=0.55\linewidth]{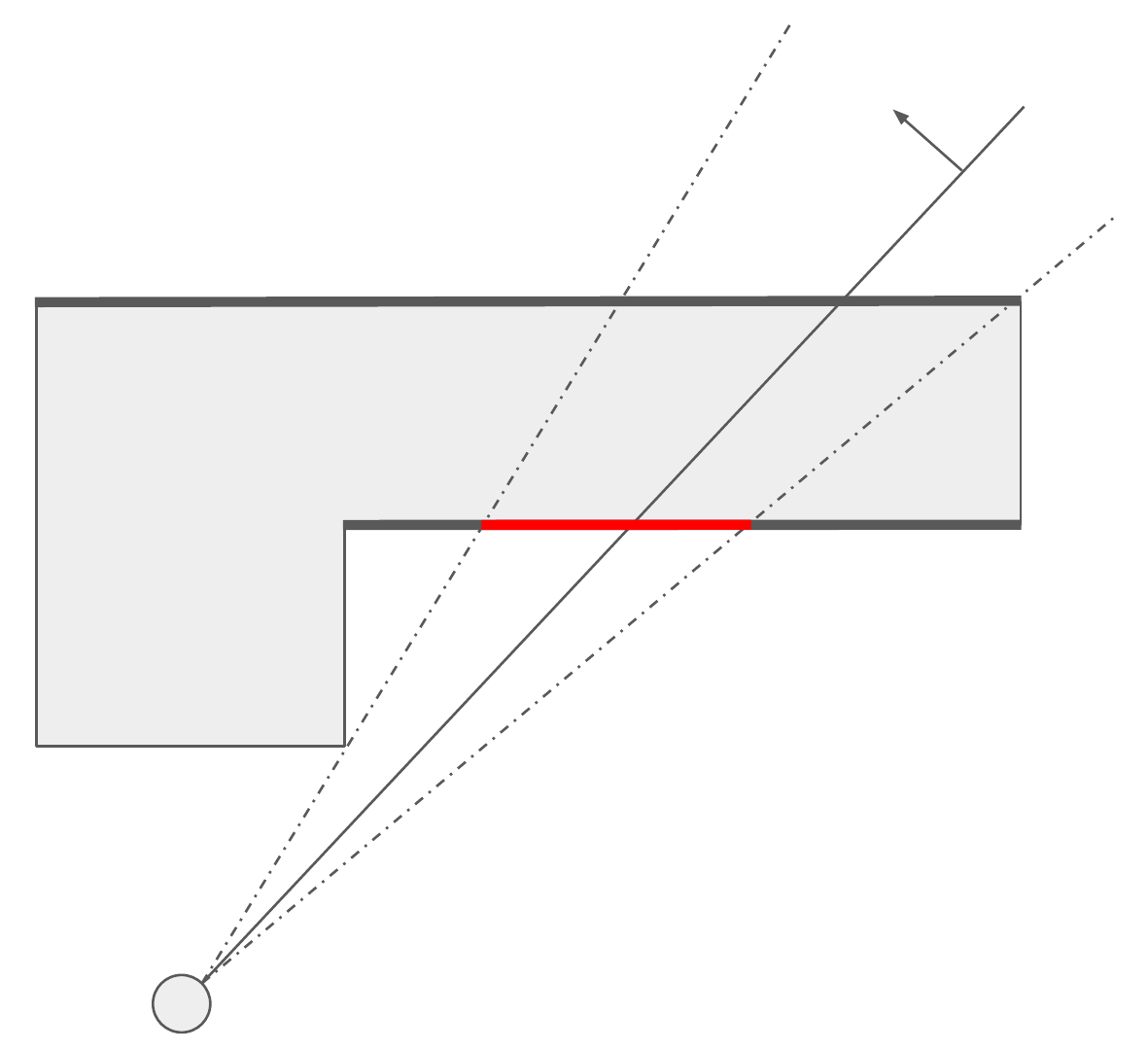}
  \caption[]{Visualization of the radial sweep algorithm. We denote a single position of the ray with a solid line, where the arrow shows the direction of movement. The last and next events are denoted with dotted lines. The current set $S$ contains the two bolded sides of the building polygon, and the bottom one is visible. In red we show the visible side limited to the current range of ray angles, which will be included in the final result. When we process the next event, two more sides will be added to $S$, and the visible building side will change.}
  \label{fig:ray_sweep_algo}
\end{figure}

The sweep line algorithm works as follows. We start with a ray originating at the street-view image location and pointing in an arbitrary direction, and rotate the ray until it makes a complete turn. Intuitively, any moment when the ray intersects the building polygon corresponds to some part of the building being visible, and the visible part is the closest intersection point of the ray with the building polygon. At any point during the sweep, we maintain the set $S$ of all polygon sides intersected by the ray. Each side needs to be inserted into $S$ exactly once during the sweep, and then (at a later point) deleted. Therefore, even though the radial sweep is a continuous process, there are only $2n$ events (each corresponding to a specific angle of the ray) that change the contents of $S$. Note that the polygon part visible at a given moment corresponds to a side in $S$ which intersects the ray closest to the origin, and this does not change as long as the set $S$ does not change. Thus, between every two consecutive events, we need to look up the closest segment in $S$, and add a part of this segment to the final result. The segment of interest can be extracted from $S$ in $O(\log n)$ time, assuming $S$ is implemented as a balanced Binary Search Tree, where the elements (polygon sides) are sorted according to their intersection with the ray. Thus, the complexity of the entire algorithm is indeed $O(n \log n)$. In Figure~\ref{fig:ray_sweep_algo} we show a single moment during the radial sweep.

\clearpage

\section{Projection pooling visualizations}

\begin{figure*}[h]
\centering
\includegraphics[width=0.48\textwidth]{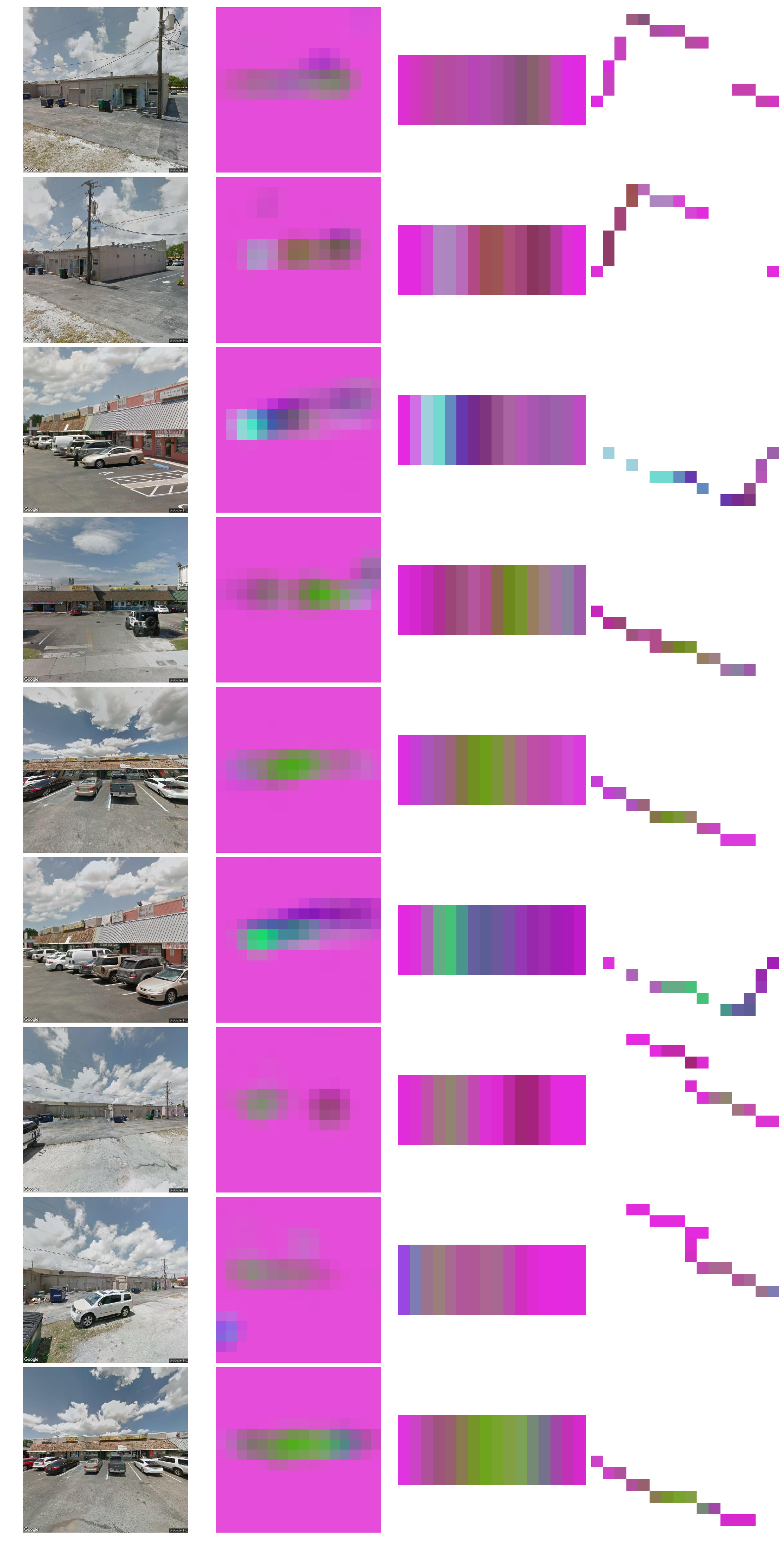}
\caption{Larger version of projection pooling visualization. Each row corresponds to one street-view image.}
\label{fig:projections_large}
\end{figure*}

\section{Infrastructure and frameworks}

We trained all our models on a Linux machine with a single GeForce GTX 1080 Ti GPU. To build our networks we used PyTorch 1.4.

\section{Hyperparameter tuning}

Here we report the hyperparameter settings that were considered during the developement of our models. We sequentially varied each of the different hyperparameters in the order reported in Table~\ref{tab:hp}.

\begin{table*}[h]
\centering
\begin{tabular}{lc}
\toprule
Hyperparameter & Values tried \\
\midrule
Fine tune batch norm & \{\textbf{No}, Yes\} \\
Crop top-view to building & \{No, \textbf{Yes}\} \\
Number of frozen ResNet blocks & \{0, 1, \textbf{2}, 3, 4\} \\
ResNet type & \{ResNet-34, \textbf{ResNet-50}\} \\
Initial learning rate & \{1e-5, 3e-5, \textbf{1e-4}, 3e-4, 1e-3, 1e-2\} \\
Weight decay & \{0.0, 1e-4, \textbf{1e-3}, 1e-2\} \\
Image augmentation & \{None, \textbf{Color jittering}, Random grayscale\} \\
\midrule
Stipe sampling strategy & \{Nearest, Sum, \textbf{Average}\} \\
Image dropout probability & \{0.0, 0.2, 0.4, \textbf{0.5}, 0.8, 0.9\} \\
Projection thickness & \{1, 2, \textbf{3}, 4\} \\
Cutout probability & \{0.0, \textbf{0.5}, 1.0\} \\
Number of street-view splits & \{1, 2, \textbf{3}, 4, 5\} \\
\bottomrule
\end{tabular}
\caption[Hyperparameter values.]{Values tried for each of the hyperparameters. Final selected value shown in bold.}
\label{tab:hp}
\end{table*}

\section{Ablation study}

\begin{table*}[h]
\centering
\begin{tabular}{cccccccccccc}
\toprule
& & & & & \multicolumn{7}{c}{Testing Accuracy} \\
\midrule
   &    &    &    &    & Constr & \#Floors & Roof & Roof & Facade & Occup &  Avg \\
SS & ID & TH & CU & SS & & & Pitch & Geom & Material &  &  \\
\midrule
N & 0   & 1 & 0   & 1 & 76 & 73.6 & 78.9 & 91.3 & 59.5 & 75.2 & \textbf{75.76} \\
S & 0   & 1 & 0   & 1 & 75.7 & 73.3 & 79 & 91.5 & 59.5 & 75.7 & \textbf{75.78} \\
A & 0   & 1 & 0   & 1 & 76.1 & 73.5 & 78.7 & 91.6 & 60 & 75.7 & \textbf{75.93} \\
A & 0.5 & 1 & 0   & 1 & 75.9 & 73.1 & 80 & 91.8 & 60.2 & 75.8 & \textbf{76.16} \\
A & 0.5 & 3 & 0   & 1 & 76.4 & 74 & 80.2 & 91.5 & 60.8 & 76.4 & \textbf{76.54} \\
A & 0.5 & 3 & 0.5 & 1 & 76.3 & 73.5 & 80.5 & 92.1 & 62.3 & 76.5 & \textbf{76.87} \\
A & 0.5 & 3 & 0.5 & 3 & 76 & 75.6 & 81.3 & 91.9 & 62.4 & 76.6 & \textbf{77.28} \\
\bottomrule
\end{tabular}
\caption[Ablation study.]{Ablation experiments for the model with projection pooling layer. SS - stripe sampling; ID - image dropout; TH - projection thickness; CU - cutout probability; SS - number of vertical splits for street-view images.}
\label{tab:ablation}
\end{table*}

\section{Impact of multiple images}

\begin{table*}[h]
\centering
\begin{tabular}{lccccccc}
\toprule
& \multicolumn{7}{c}{Testing Accuracy} \\
\midrule
& Constr & \#Floors & Roof  & Roof & Facade   & Occup & Avg \\
&        &          & Pitch & Geom & Material &       &     \\
\midrule
1xSV + TV       & 75.7 & 72.0 & 80.3 & 91.0 & 60.3 & 74.4 & \textbf{75.60} \\
Up to 2xSV + TV & 76.3 & 74.4 & 81.1 & 92.1 & 61.9 & 75.9 & \textbf{76.96} \\
Up to 3xSV + TV & 76.1 & 75.2 & 81.4 & 91.9 & 62.3 & 76.4 & \textbf{77.22} \\
Up to 4xSV + TV & 76.1 & 75.5 & 81.1 & 92.1 & 62.5 & 76.3 & \textbf{77.26} \\
All SV + TV & 76.0 & 75.6 & 81.3 & 91.9 & 62.4 & 76.6 & \textbf{77.28} \\
\bottomrule
\end{tabular}
\caption[Impact of the maximum number of street-view images per example.]{Impact of the maximum number of street-view images per example on model performance.}
\label{tab:results_multiple}
\end{table*}

\clearpage

\bibliography{references}